\pdfoutput=1
\documentclass[11pt]{article}
\usepackage[final]{acl}
\usepackage{times}
\usepackage{latexsym}
\usepackage[T1]{fontenc}
\usepackage[utf8]{inputenc}
\usepackage{microtype}
\usepackage{inconsolata}
\usepackage{graphicx}
\usepackage{soul}
\usepackage{xcolor}
\usepackage{placeins}
\usepackage{amsmath}
\usepackage{amssymb}
\usepackage{makecell}
\usepackage{booktabs}
\usepackage{multicol}
\usepackage{multirow}
\usepackage{arydshln}
\usepackage{cleveref}
\usepackage{enumitem}
\setlist[itemize]{itemsep=0.5ex,topsep=0.5ex}
\title{Is It Good Data for Multilingual Instruction Tuning or \\ Just Bad Multilingual Evaluation for Large Language Models?}

\author{
Pinzhen Chen\textsuperscript{1} \quad Simon Yu\textsuperscript{1,2} \quad Zhicheng Guo\textsuperscript{3} \quad Barry Haddow\textsuperscript{1}\\
\textsuperscript{1}University of Edinburgh \quad \textsuperscript{2}Northeastern University \quad \textsuperscript{3}Tsinghua University\\
\texttt{pinzhen.chen@ed.ac.uk \quad yu.chi@northeastern.edu}\\
\texttt{guo-zc21@mails.tsinghua.edu.cn \quad bhaddow@ed.ac.uk}
}

\begin{document}
\maketitle
\begin{abstract}
Multilingual large language models are designed, claimed, and expected to cater to speakers of varied languages. We hypothesise that the current practices of fine-tuning and evaluating these models may not perfectly align with this objective owing to a heavy reliance on translation, which cannot cover language-specific knowledge but can introduce translation defects. It remains unknown whether the nature of the instruction data has an impact on the model output; conversely, it is questionable whether translated test sets can capture such nuances. Due to the often coupled practices of using translated data in both stages, such imperfections could have been overlooked. This work investigates these issues using controlled native or translated data during the instruction tuning and evaluation stages. We show that native or generation benchmarks reveal a notable difference between native and translated instruction data especially when model performance is high, whereas other types of test sets cannot. The comparison between round-trip and single-pass translations reflects the importance of knowledge from language-native resources. Finally, we demonstrate that regularization is beneficial to bridging this gap on structured but not generative tasks.\footnote{Our code and data will be made public at \url{https://github.com/PinzhenChen/good-data-or-bad-eval}.}
\end{abstract}

\section{Introduction}
Instruction tuning, or supervised fine-tuning, can prepare a large language model (LLM) for better task generalization and natural interactions in downstream applications \citep{mishra-etal-2022-cross,wei2022finetuned,sanh2022multitask,ouyang_rlhf}. Major efforts of building instruction datasets centre on English \citep{wei2022finetuned, alpaca, DatabricksBlog2023DollyV2, Ivison2023CamelsIA}, whereas the multilingual counterparts remain modest in size, variety, and coverage. Many multilingual instruction datasets have been seeded from English data and developed using translation as part of the pipeline \citep{li2023bactrian, Chen_MultilingualSIFT, chen-etal-2024-monolingual, lai-etal-2023-okapi}. A notable exception is Aya, a year-long project that invited volunteers around the globe to write and edit prompt-response examples in their native language \citep[][]{singh2024aya}, making it a language-native dataset.\footnote{\hspace{0.25ex}``Aya Dataset'' but not ``Aya Collection'' which comprises many translated components.}

Although the Aya dataset was contributed by volunteers, it still carries a high social utility cost considering the personnel hours devoted. In contrast, translating existing resources by machine, or even by human, is a more convenient option. Nonetheless, translated data carry imperfections \citep{clark-etal-2020-tydi, artetxe-etal-2020-translation}: 1) it represents the culture and knowledge specific to the original language; 2) the translation process introduces translationese, an unnatural language style \citep{gellerstam, baker-corpus}, as well as errors since certain content or tasks can be corrupted, e.g. grammatical error correction. On the other hand, recent research discovered that instruction tuning is ``superficial'' where an LLM mainly learns the response format \citep{zhou2023lima}, and it cannot enhance knowledge at the current scale \citep{ghosh2024closer}. These insights imply that the shortcomings of translated data might not propagate into an instruction-tuned model. Hence, when widening language support, an important question arises: \emph{Is translated data sufficient for instruction tuning?}

We regard ``sufficiency'' as the fact that, when used to fine-tune an LLM, translated instructions should lead to output quality similar to that of native data. Yet, examining this training data factor cannot be separated from carefully considering the evaluation protocol, because many existing multilingual benchmarks have been created via translation. This translation bias, if present in both training and test sets, could hinder a meaningful conclusion. We thus put forward our second research question: \emph{If translated and native instruction data make a difference, would a translated benchmark capture it?} Subsequently, we use round-trip translated data to answer: \emph{Which is the cause of the gap, translation defects or missing language-specific knowledge in instructions?} Finally, when translated data is hardly avoidable: \emph{What techniques can we adopt to bridge the performance gap?}

This work systematically investigates native and translated data used during instruction tuning and evaluation. We experiment with eight models of varying sizes and data distributions and evaluate these models on nine benchmarks of different natures: translated versus native as well as classification versus generation. Empirical results suggest that a prudent choice in multilingual LLM evaluation is crucial. Foreshadowing the answers to the research questions raised earlier:
\begin{enumerate}
  \item Native and translated data can lead to a performance gap on several benchmarks, especially when the model performance is strong.
  \item Such a difference is more pronounced on benchmarks that are natively created (TyDi QA, CMMLU, C-Eval) or generative in nature (XQuAD, open-ended QA) compared to translated structured tests (MT/HT-MMLU).
  \item Round-trip translation from native data outperforms single-pass translation from English data, implying that missing language-specific knowledge could be more detrimental than having translation defects.
  \item Regularization during instruction tuning time, e.g. using a lower learning rate or multilingual instruction tuning, can be beneficial if translated data has to be used. It can close the native-translated performance gap on structured tasks but not generative tasks.
\end{enumerate}
These insights mean that opposite conclusions can be made when different combinations of instruction and test sets are adopted. Based on the findings, we recommend multilingual LLMs be evaluated on a range of benchmarks to include language-native and generative tasks.

\section{Related Work}

\subsection{Instruction tuning data}
Instruction data can be created by writing questions and responses from scratch \citep{DatabricksBlog2023DollyV2,singh2024aya}, collecting user-system interactions \citep{kopf2023openassistant}, or templating structured data instances into natural texts \citep{mishra-etal-2022-cross, sanh2022multitask}. It is also feasible to distil large language models by feeding existing examples \citep{alpaca,wei2023polylm}. Stemming from English data, many multilingual instruction datasets, especially open-ended question-response pairs, have been created via machine translation \citep{muennighoff-etal-2023-crosslingual,chen2023breaking, Chen_MultilingualSIFT, chen-etal-2024-monolingual, chai2024xcot, lai2024mcot}. Slightly differently, \citet{li2023bactrian} translated English questions into multiple languages but used GPT to generate responses to avoid translationese. These options are more affordable than creating language-native data directly, but they are not flawless since they can introduce generation errors and knowledge-language mismatches.

In recent research progress on LLM instruction tuning, the ``superficial alignment hypothesis'' \citep{zhou2023lima} might offer some relief to these concerns. It claims that a strong foundation model mostly learns the response template from instruction tuning---therefore the translation artefacts or language-specific knowledge would not be overly consumed \citep{ghosh2024closer}. To our knowledge, there is no prior work that systematically compared native and translated instruction data.

\subsection{Multilingual LLM evaluation}
Machine translation has been used to extend several benchmarks to more languages~\citep[][and the list is growing]{conneau-etal-2018-xnli, artetxe-etal-2020-cross, dumitrescu2021liro, bandarkar2023belebele}. Many studies exploring multilingualism in LLMs yielded findings based on translated instruction data and/or translated evaluation sets, from the earlier mT5 to the concurrent Llama 3.1 \citep[][]{Xue2020mT5AM, Caete2023SpanishPB, Ahuja2023MEGAME, Chinese-llama-alpaca, puduppully-etal-2023-decomt, yang2023bigtranslate, lai-etal-2023-okapi, kew2023turning, chen-etal-2024-monolingual, singh2024aya, ji2024lucky, liu2024translation, shaham2024multilingual,dubey2024llama}. While these works have significantly pushed the boundary of multilingualism in LLMs, we attempt to revisit the effect of using translated data.

\citet{clark-etal-2020-tydi} discussed the disadvantages of using translated tests: they incorporate translationese and represent the source language's knowledge; \citet{artetxe-etal-2020-translation} revealed how minor translation artefacts can significantly impact evaluation outcomes. It has been shown and argued that, albeit intuitively, translated training data improves scores on test data created via translation \citep{singh2019xlda, artetxe-etal-2020-translation}. The machine translation community found that translated test input ``can have a detrimental
effect on the accuracy of evaluation'' \citep{Lubli2020-a-set, graham-etal-2020-statistical, akhbardeh-etal-2021-findings}.  This paper demonstrates that by altering the nature of the instruction or evaluation data, evaluation can lead to different conclusions for LLM instruction tuning.

Our comparative analysis of native and translated data also relates to understanding the integrity of LLM evaluation and the representation of language-specific knowledge from a meta-evaluation perspective. It is the expectation of the users that an LLM should not merely exhibit linguistic fluency but also embed the culture tied to the languages. We believe this to be a crucial and timely topic in the current LLM landscape. Earlier, \citet{lyu2024beyond} examined various mechanisms of obtaining LLM responses. Concurrently, \citet{gema2024mmlu} found correctness issues in a specific benchmark; \citet{etxaniz2024bertaqa} showed that models can have distinct behaviours on local and global knowledge; \citet{gu2024olmes} called for transparency in choosing formatting and configurations. In comparison, our work looks at multilingual evaluation from the dimension of data characteristics.

\section{Experiment Design}

\subsection{Instruction data}\label{sec:instruction_data}
The focus of our study is to answer the research questions on the nature of instruction data and evaluation data as well as their impact on a trustworthy evaluation. We experiment with non-English training and test data created through distinct procedures: \textbf{created natively} and \textbf{translated}. We run monolingual instruction tuning: an LLM is fine-tuned in a single language every time to prevent potential cross-lingual influences.

\paragraph{Languages}
We study model performance in three languages---Spanish (es), Russian (ru), and Chinese (zh)---with the following considerations: 1) these languages cover a combination of different language families and writing scripts; 2) they are medium-to-high resourced, where the quality of the data, native or translated instructions, is satisfactory; 3) their presence in LLM pre-training data is significant, so we can expect reasonable output quality.

\paragraph{Native data}
We use the training split in the Aya dataset \citep{singh2024aya}, which was written from scratch and then edited by human annotators in their native language. The Spanish, Russian, and Chinese training sets have a size of 3854, 423, and 3944 each.

\paragraph{Translated data}
We generate translated data equivalent in volume to the native data. This is done by sampling Aya's English split to match the size of native data in each language and translating the sample to that language. We always translate the instructions and responses separately. Two distinct versions of translated data are obtained via Google Translate and Cohere Command R\footnote{\url{https://docs.cohere.com/docs/command-r}}. Google Translate is a well-known commercial translation engine, whereas Command R, a large language model, is capable of adhering to more customised guidelines. Technically, we prompt Command R to maintain the original data formatting while translating, as illustrated \Cref{app:translation-prompt}.

\subsection{Close-ended evaluation}
We perform automatic evaluations on close-ended tasks, where a model is expected to generate a pre-defined response given a question. The evaluation covers multilingual understanding and reasoning tasks commonly used to benchmark LLMs. These test sets come from various sources such as native annotation, human translation, and machine translation. All evaluations are conducted using \texttt{lm-evaluation-harness} \citep{lm-eval-harness} with default settings unless stated below.

\paragraph{Native benchmarks} We first evaluate our instruction-tuned models on test sets that have been constructed from scratch by native speakers, on which we hypothesize a performance gap between native and translated instruction fine-tuning. 
\begin{itemize}
    \item \textbf{TyDi QA}: created by inviting native speakers to write down questions related to articles shown to them \citep{clark-etal-2020-tydi}. We use its Russian split. We run 1-shot prompting and measure models' F1 scores. 
    \item \textbf{CMMLU} \citep{Li2023CMMLUMM} and \textbf{C-Eval} \citep{huang2024c}: both are multi-disciplinary tests containing questions on the Chinese culture and domain, made from resources in Chinese. We prompt with 5-shot examples and use accuracy as the metric. 
\end{itemize}
Unfortunately, we could not identify a native benchmark that assesses general knowledge in Spanish. 

\paragraph{Translated benchmarks} We use four translated benchmarks including both human-translated and machine-translated test sets. Most of these cover the three languages we study.
\begin{itemize}
    \item \textbf{XQuAD}: a question answering task requiring text extraction from a given context \citep{artetxe-etal-2020-cross}, human-translated from the English SQuAD \citep{rajpurkar-etal-2016-squad}. Evaluation is done in a 0-shot setting. We adopt two metrics: a strict string-level exact match (EM) and a lenient ``include'' checking whether the reference is a substring of the model generation.
    \item \textbf{MGSM}: grade school mathematics questions human-translated from the English GSM8K \citep{cobbe2021training,Shi2022LanguageMA}. We provide 5-shot examples with chain-of-thought and measure exact token match.
    \item \textbf{MT-MMLU}: \citet{lai-etal-2023-okapi}'s ChatGPT-translated multilingual MMLU \citep{hendryckstest2021}, designated as MT-MMLU in our work. We use 5-shot prompting and accuracy as the metric.
    \item \textbf{HT-MMLU}: a professionally human-translated (HT) edition\footnote{\url{https://huggingface.co/datasets/openai/MMMLU}} of MMLU released when our camera-ready paper is being prepared. \Cref{sec:cause} offers a preliminary study of model behaviours on HT-MMLU and MT-MMLU to understand the impact of human and machine translation.
\end{itemize}

\subsection{Open-ended generation}
We then evaluate models with open-ended question answering (QA) under controlled translated and native settings:
\begin{itemize}
    \item \textbf{Translated}: 50 English questions from OpenAssistant \citep[OASST1;][]{kopf2023openassistant} and then human-translated by \citet{chen-etal-2024-monolingual}. We use the translated questions in Spanish, Russian, and Chinese.
    \item \textbf{Native}: 50 questions in Spanish, Russian, and Chinese, directly sampled from OASST1. We only use the first-round queries in multi-turn conversations.
\end{itemize}
Given the open-ended nature, there is no gold response to compare a model generation against. To avoid expensive human evaluation at scale, we use LLM-as-a-judge, which has shown a strong correlation with human judgement \citep{zheng2023judging}. We use two LLM judges other than the translators to avoid LLM preference bias: GPT-4-Turbo and Command R+.\footnote{Both accessed via API in Apr 2024.} The judges directly score each instruction-response pair according to a 5-point Likert scale (1 to 5), which can avoid position bias in response comparison. The total score for a model therefore ranges between 50 to 250. The exact wording of the judging prompt is the same for both LLMs and is attached in \Cref{app:llm-as-a-judge-prompt} \Cref{fig:llm-eval-instruction}.

\section{Experiments and Analysis}
\subsection{Technical setup}

\paragraph{Base models}
We fine-tune base models of different sizes from three sources: 1) Llama 2 at 7B, trained on 2T tokens with a 32K vocabulary and released in Jul 2023 \citep{llama2}; 2) Gemma at 2B and 7B (circa 8.54B parameters), trained on 3T and 6T tokens respectively with a 256K vocabulary and released in Feb 2024 \citep{gemma}; 3) Qwen 1.5 at 0.5B, 1.8B, 4B, 7B, and 14B released in Feb 2024 \citep{qwen1.5}.\footnote{All models were up-to-date when the experiments were conducted in Apr 2024 but became one generation behind by the time the paper was accepted in Sep 2024.}

\paragraph{Instruction tuning}

Let $I$ represent an instruction and $Y=y_1,y_2,...,y_{|Y|}$ a sequence of output tokens. The instruction is first templated into a pre-defined format, denoted as $\mathcal{T}(I)$. We fine-tune an LLM parameterised by $\theta$ by optimising the log-likelihood on the output tokens only: $\mathcal{L}(Y,\mathcal{T}(I);\theta)=-\log P(Y|\mathcal{T}(I); \theta)$.

We apply low-rank adaptation where the base model is loaded in 8-bit and frozen during training \citep{hu2022lora,dettmers2023qlora}. We attach to all key, query, and value matrices a low-rank adapter with a rank of 8, an alpha of 16, and a dropout of 0.05. The learning rate is set to $10^{-4}$ and the effective batch size to 64. All models are given a training budget of 10 epochs and we validate perplexity on held-out instruction data after each epoch to keep the best checkpoint. We used a combination of NVIDIA 3090-24G, A100-40G, and A100-80G GPUs. Fine-tuning took 1 to 7 hours depending on the model and data size.

\begin{figure}[t]
  \centering\small
  \input{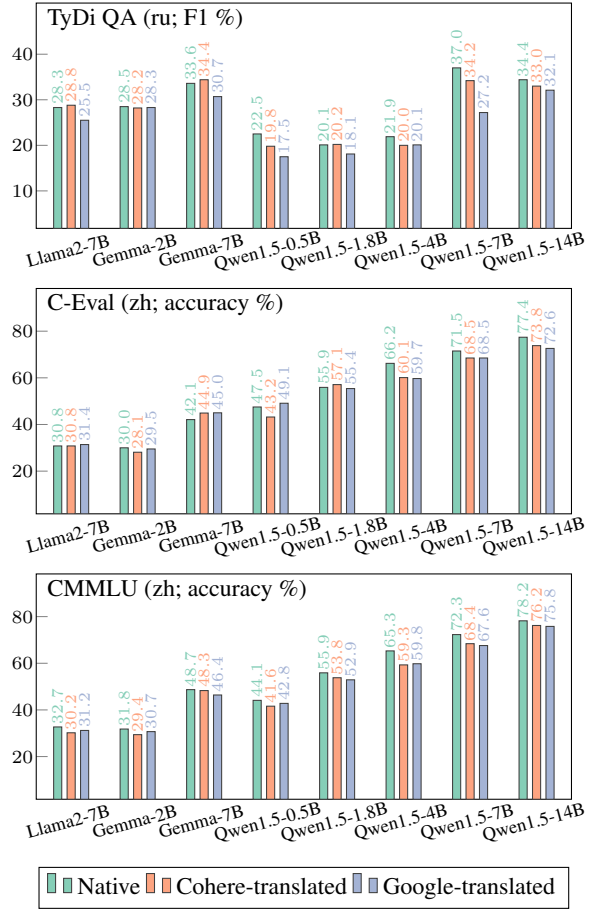}
  \vspace{-3.5ex}
  \caption{Results on {native close-ended test sets}: native instruction-tuned models have an edge.}
  \vspace{-2ex}
  \label{fig:native-close-ended-tests}
\end{figure}

\begin{figure}[t]
  \centering\small
  \input{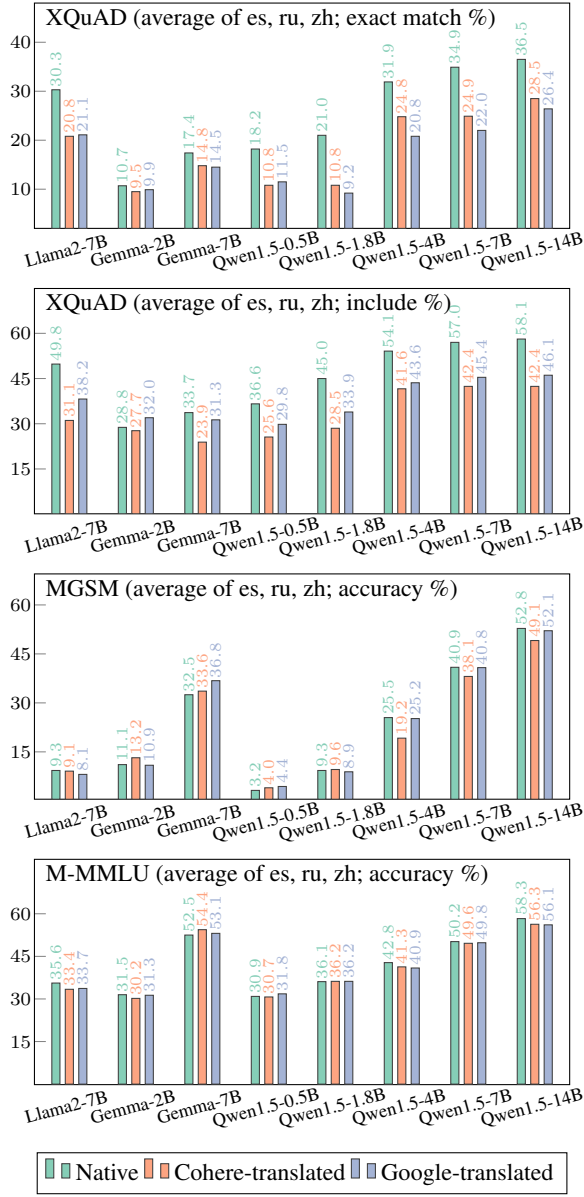}
  \vspace{-3.5ex}
  \caption{Results on {translated close-ended test sets}: native instruction-tuned models are superior on XQuAD, but all data conditions have comparable results on MGSM and MT-MMLU.}
  \vspace{-2ex}
  \label{fig:translated-close-ended-tests}
\end{figure}

\begin{figure}[t]
  \centering\small
  \input{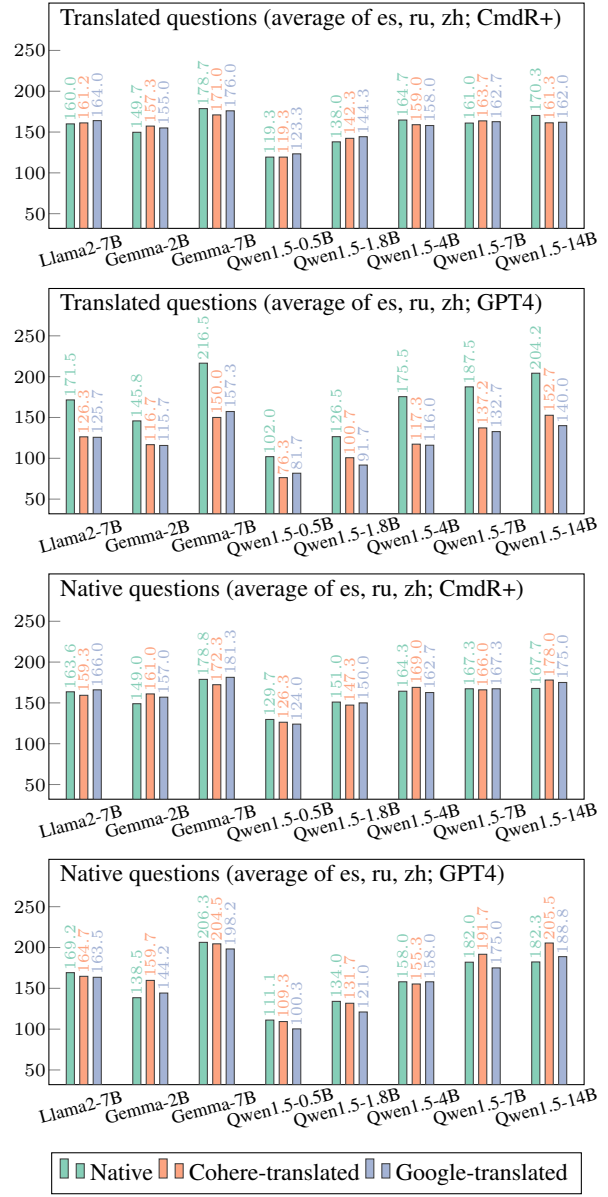}
  \vspace{-3.5ex}
  \caption{Results on native and translated open-ended question answering: native instruction-tuned models are superior for translated questions when judged by GPT-4-Turbo, but all data conditions result in similar numbers in other cases.}
  \vspace{-2ex}
  \label{fig:open-ended-tests}
\end{figure}

\subsection{Is there a gap, and on what?}\label{sec:gap}
We display results for the native tests, TyDi QA, C-Eval, and CMMLU, in \Cref{fig:native-close-ended-tests}. It shows that models fine-tuned with native instructions surpass those fine-tuned with translated data in most cases with consistent patterns across the two languages.

In terms of translated multilingual benchmarks, \Cref{fig:translated-close-ended-tests} exhibits diverging trends. On the XQuAD benchmark, using native instruction data consistently and significantly outperforms translated data under both metrics, however, it loses the advantage on MGSM and MT-MMLU.

For open-ended QA, we show different combinations of the test data (native or human-translated) and judges (GPT-4-Turbo or Command R+) in \Cref{fig:open-ended-tests}. The largest native-translated discrepancy occurs when models are tested on translated questions and judged by GPT-4-Turbo. When testing on translated questions and judged by Command R+, native data is slightly ahead when the model size is big. In other cases, native data is not better than translated data. These results also suggest that the LLM-as-a-judge metric affects empirical results too. However, it is difficult to arrive at a clear conclusion since we do not have transparent access to the data used in GPT or Command models---it might be the case that these models have been instruction-tuned with translated data.

Overall, we see that in terms of model performance, native data can surpass translated data under some evaluations, which suggests that translated instruction data is not always sufficient. While these observations have been made from the aspect of data/model performance, they cannot be decoupled from the potential test set imperfections. Assuming native data should lead to better metric numbers, it has been revealed that two types of evaluation benchmarks are more effective in reflecting this: 1) those that originate in the test language itself (TyDi QA, C-Eval, and CMMLU) and 2) those that are generative in nature (XQuAD and open-ended questions) even though they could have been translated from English.

\subsection{When is the gap obvious?}\label{sec:obvious}

We hypothesise that the output quality difference between using native and translated data would be more noticeable when a model's overall performance is better---namely, the subtle translation bias might not be pivotal if a model's capability is low enough that many instances are incorrectly predicted in the first place. Hence, for each previous benchmark where native data outperforms translated data, we run a post hoc analysis on the correlation between the native data performance and the native-translated performance gap.

We average the {Cohere}-translated and {Google}-translated scores to represent the final score for translated data. The score difference between models instruction-tuned on native data and translated data can then be defined as $\Delta S = S_\text{native} - \frac{1}{2} (S_\text{cohere} + S_\text{google})$, where $S_\text{native}$, $S_\text{cohere}$, and $S_\text{google}$ stand for model scores on native, Cohere-translated, and Google-translated data respectively. Then, we compute the Pearson correlation coefficient $r_{\Delta S, S_\text{native}}$ between $\Delta S$ and $S_\text{native}$ for each test set. It is worth noting that we consider all individual languages' scores instead of the averaged number across languages where applicable.

\begin{figure}[t]
  \centering\small
  \input{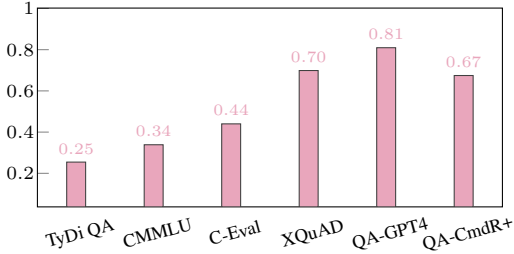}
  \vspace{-2ex}
  \caption{Pearson's correlation between \textit{native data performance} and \textit{native-translated performance difference} for various benchmarks: weaker correlation for structured tasks and stronger correlation for generative tasks.}
  \vspace{-3ex}
  \label{fig:abs_diff_correlation}
\end{figure}

We cover all benchmarks where a clear native-translated difference has been observed earlier. Open-ended question answering is abbreviated as QA-GPT4 and QA-CmdR+ depending on the LLM judge used. The outcome is shown in \Cref{fig:abs_diff_correlation}: the correlation between $\Delta S$ and $S_\text{native}$ is weak for structured tasks like TyDi QA, C-MMLU, and C-Eval, but very strong for tasks involving generation like XQuAD and open-ended QA. This pattern indicates that 1) concerning the instruction data, the nature of being native or translated shines through as the model performance gets higher; 2) on the evaluation end, such data difference leaves a more pronounced gap on generative benchmarks. On a related note, in \citet{kew2023turning}'s study, it is shown that cross-lingual transfer is more prominent in generative tasks but less in classification tasks. Altogether, it might be conjectured that the instruction data quality plays a more crucial role when a model is evaluated by generative tasks as opposed to classification tasks.

\begin{figure}[t]
    \centering\small
    \resizebox{1\linewidth}{!}{%
    \input{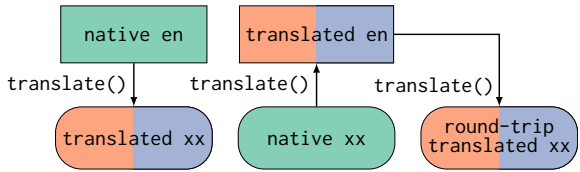}
    }
    \vspace{-2ex}
    \caption{Round-trip translation (via English) produces translated data sharing the same origin as native data.}
    \vspace{-3ex}
    \label{fig:round-trip-translation}
\end{figure}

\subsection{What potentially causes the gap?}\label{sec:cause}
\paragraph{Knowledge mismatch vs translation defects in instructions} Translating instruction data introduces these imperfections. To understand which accounts more for model degradation, we disentangle the two elements in instructions using round-trip translation (RTT): we translate native data from one language into another and then translate it back to the original language, as illustrated in \Cref{fig:round-trip-translation}. By doing so, we can have a ``translated'' dataset that preserves the same knowledge and domain as the original data but contains translation defects.

We construct the RTT version of Russian and Chinese instruction data from their native data with Cohere or Google translation pivoting via English. This follows the same procedure used to obtain translated instruction data in \Cref{sec:instruction_data}, except that the translation workflow is now done twice: X$\rightarrow$English followed by English$\rightarrow$X. We then compare models trained on RTT data with those trained on data translated from English on native benchmarks (TyDi QA and CMMLU).

\begin{table}[t]
    \centering\small
    \setlength{\tabcolsep}{0.8ex}
    \begin{tabular}{lccccc}
\toprule
        \multicolumn{1}{c}{\multirow[b]{3}{*}{Base Model}} & \multicolumn{2}{c}{Cohere} & \multicolumn{2}{c}{Google} \\
        \cmidrule(lr){2-3}\cmidrule(lr){4-5}
        & \makecell{RTT\\ru-origin} & \makecell{translated\\en-origin} & \makecell{RTT\\ru-origin} & \makecell{translated\\en-origin} \\
\midrule
        Llama2-7B   & 25.9 & \textbf{28.8} & \textbf{25.7} & 25.5 \\ 
        Gemma-7B    & 29.4 & \textbf{34.4} & \textbf{33.3} & 30.7 \\
        Qwen1.5-4B  & \textbf{23.0} & 20.0 & \textbf{22.4} & 20.1 \\
        Qwen1.5-7B  & \textbf{35.5} & 34.2 & \textbf{34.9} & 27.2 \\
        Qwen1.5-14B & 30.4 & \textbf{33.0} & 30.7 & \textbf{32.1} \\
\bottomrule
    \end{tabular}
    \caption{Results for models trained on RTT data (ru-origin) or translated data (en-origin) on TyDi QA (ru).}
    \label{tab:rtt_tydiqa}
\vspace{3ex}
\centering\small
    \setlength{\tabcolsep}{0.8ex}
    \begin{tabular}{lccccc}
\toprule
        \multicolumn{1}{c}{\multirow[b]{3}{*}{Base Model}} & \multicolumn{2}{c}{Cohere} & \multicolumn{2}{c}{Google} \\
        \cmidrule(lr){2-3}\cmidrule(lr){4-5}
        & \makecell{RTT\\zh-origin} & \makecell{translated\\en-origin} & \makecell{RTT\\zh-origin} & \makecell{translated\\en-origin} \\
\midrule
        Llama2-7B   & \textbf{31.6} & 30.2 & \textbf{32.2} & 31.2 \\ 
        Gemma-7B    & \textbf{48.6} & 48.3 & \textbf{48.4} & 46.4 \\
        Qwen1.5-4B  & \textbf{63.7} & 59.3 & \textbf{64.6} & 59.8 \\
        Qwen1.5-7B  & \textbf{68.9} & 68.4 & \textbf{70.5} & 67.6 \\
        Qwen1.5-14B & \textbf{77.5} & 76.2 & \textbf{77.4} & 75.8 \\
\bottomrule
    \end{tabular}
    \caption{Results for models trained on RTT data (zh-origin) or translated data (en-origin) on CMMLU (zh).}
    \label{tab:rtt_cmmlu}
\vspace{3ex}
\centering\small
    \setlength{\tabcolsep}{0.6ex}
    \begin{tabular}{llccccc}
\toprule
\multicolumn{1}{c}{\multirow[b]{3}{*}{Base Model}} & \multicolumn{1}{c}{\multirow[b]{3}{*}{Data}} & \multicolumn{2}{c}{Spanish} & \multicolumn{2}{c}{Chinese} \\
\cmidrule(lr){3-4}\cmidrule(lr){5-6}
& & \makecell{MT-\\MMLU} & \makecell{HT-\\MMLU} & \makecell{MT-\\MMLU} & \makecell{HT-\\MMLU} \\
\midrule
\multirow{3}{*}{Llama2-7B}
& native & 38.4 & 37.6 & 34.4 & 33.8 \\ 
& cohere & 38.0 & 37.2 & 27.6 & 27.8 \\
& google & 36.4 & 35.9 & 30.4 & 29.5 \\
\cdashlinelr{1-6}
\multirow{3}{*}{Gemma-7B}
& native & 55.9 & 54.9 & 48.7 & 48.0 \\ 
& cohere & 58.4 & 57.5 & 50.8 & 50.3 \\ 
& google & 56.1 & 55.6 & 49.7 & 48.8 \\ 
\cdashlinelr{1-6}
\multirow{3}{*}{Qwen1.5-4B}
& native & 40.9 & 40.2 & 49.3 & 49.0 \\ 
& cohere & 39.9 & 39.0 & 44.5 & 45.2 \\ 
& google & 39.6 & 39.0 & 44.2 & 45.0 \\ 
\cdashlinelr{1-6}
\multirow{3}{*}{Qwen1.5-7B}
& native & 50.3 & 49.6 & 52.9 & 53.0 \\ 
& cohere & 49.6 & 48.4 & 52.6 & 51.8 \\ 
& google & 50.4 & 49.3 & 51.8 & 51.3 \\ 
\cdashlinelr{1-6}
\multirow{3}{*}{Qwen1.5-14B}
& native & 58.1 & 57.8 & 61.3 & 60.7 \\ 
& cohere & 55.8 & 55.1 & 57.9 & 57.7 \\ 
& google & 54.9 & 54.2 & 58.0 & 57.3 \\ 
\bottomrule
    \end{tabular}
    \caption{Results for models trained on different data on MT-MMLU and HT-MMLU.}
    \vspace{-1ex}
    \label{tab:mt-mmlu_vs_ht-mmlu}
\end{table}

Regarding TyDi QA in \Cref{tab:rtt_tydiqa}, we notice mixed results for Cohere translation but a relative advantage in RTT with Google translation. For CMMLU in \Cref{tab:rtt_cmmlu}, models with RTT (test language-origin) are uniformly better than those with data translated from English. RTT's strong performance---despite having undergone the translation process twice which likely leaves more translationese and errors---signifies the importance of incorporating native knowledge when widening language support in multilingual language models.

\paragraph{Human vs machine translated test sets}
Comparing MT-MMLU and HT-MMLU results can reveal the impact of human and machine translation on the evaluation end. This comparison is carefully controlled where both test sets have the same questions originating in English and testing the same knowledge. During testing, the same set of demonstrations is prepared for the same question across the two tests. We list Spanish and Chinese results in \Cref{tab:mt-mmlu_vs_ht-mmlu} which are very similar on the two benchmarks and the native-translated gap is smaller compared with those on native or generative tasks. As shown in \Cref{app:comprehensive_results} \Cref{tab:mmmlu-1e-6,tab:openai_mmlu-1e-6}, the gaps even disappear under a lower learning rate.

Interestingly, gap patterns are consistent across the two translated MMLU tests: Llama2-7B on Chinese, Qwen1.4-4B on Chinese, and Qwen1.5-14B on both languages. These observations imply that both test sets are homogeneous and that (good) MT can match professional HT in expanding test set language coverage. This also corroborates our early finding that missing language-specific knowledge can be a more differentiating factor.

\subsection{Can we bridge the gap?}
Despite \Cref{sec:cause} suggesting that it is more critical to have the domain and the knowledge of the native language in instructions, it is an unrealistic setting since it employs native data. This is difficult to obtain especially for under-served languages, so it is hard to avoid machine-translated data. We, therefore, investigate techniques that can apply better regularization during instruction tuning to reduce the negative impact of the translated data. This also represents an effort to pursue a more generalizable finding.

\paragraph{A lower learning rate} Our first inspiration is drawn from \citet{chirkova2024zeroshot}, whose experiments showed that English instruction-tuned models display remarkably different levels of cross-lingual transfer when only changing the learning rate---a smaller one leads to better instruction following in zero-shot languages. This means that it is possible to teach a base model a desired instruction-response style without even touching on the content or language. In this case, the undesirable properties in translated data could be mitigated. Following this, we run another set of experiments with the learning rate reduced from $10^{-4}$ to $10^{-6}$.

\paragraph{Multilingualism}

Another exploration is multilingual instruction tuning, which could prevent a model from overfitting to a single language. In addition to Spanish, Russian, and Chinese which we evaluate, we also add another five languages---Arabic (ar), German (de), Finnish (fi), Irish (ga), and Hindi (hi)---into the multilingual pot. For the native multilingual data, we simply down-sample all languages in the Aya dataset to a size of 241 (the size of the German split in Aya, which is the smallest among the eight languages), leading to a total size of 1928. For the translated data in each language, we randomly select 241 instances from English and translate them (different data instances for different languages). This simulates a multilingual instruction set derived from translating English resources.

\begin{table}[t]
\setlength{\tabcolsep}{4.5pt}
\centering\small
\begin{tabular}{llccc}
\toprule
\multicolumn{1}{c}{\multirow[c]{2}{*}{Base Model}}  & \multicolumn{1}{c}{\multirow[c]{2}{*}{Data}} & $10^{-6}$$\leftarrow$ & $10^{-4}$ & $10^{-4}$ \\
  &  & Mono & Mono & $\rightarrow$Multi\\
\midrule
  &  native &  \textbf{33.4} &  28.3 &  25.1\\
Llama2-7B &  cohere &  \ul{33.3} &  28.8 &  23.4 \\
  &  google &  \ul{33.3} &  25.5 &  22.9 \\
\cdashlinelr{1-5}
  &  native &  \textbf{37.7} &  33.6 &  31.5 \\
Gemma-7B &  cohere &  \ul{38.1} &  34.4 &  31.4 \\
  &  google &  \ul{37.9} &  30.7&  30.9 \\
\cdashlinelr{1-5}
  &  native &  22.4 & \textbf{37.0}  &  \textbf{37.0} \\
Qwen1.5-7B &  cohere &  22.9 &  34.2&  33.0 \\
  &  google &  22.7 &  27.2&  27.1 \\
\cdashlinelr{1-5}
  &  native &  24.8 &   \textbf{34.4} &  32.8 \\
Qwen1.5-14B &  cohere &  24.6 &  33.0 &  29.3 \\
  &  google &  24.9 &  32.1&  \ul{35.2} \\
\bottomrule
\end{tabular}
\vspace{-1ex}
\caption{Sometimes the gap can be closed on TyDi QA.}
\end{table}

\begin{table}[t]
\setlength{\tabcolsep}{4.5pt}
\centering\small
\begin{tabular}{llccc}
\toprule
\multicolumn{1}{c}{\multirow[c]{2}{*}{Base Model}}  & \multicolumn{1}{c}{\multirow[c]{2}{*}{Data}} & $10^{-6}$$\leftarrow$ & $10^{-4}$ & $10^{-4}$ \\
  &  & Mono & Mono & $\rightarrow$Multi\\
\midrule
  &  native & 31.8 & \textbf{32.7} & \textbf{32.6} \\
Llama2-7B &  cohere & 32.0 & 30.2 & \ul{32.7} \\
  &  google & 32.0 & 31.2 & 32.1 \\
\cdashlinelr{1-5}
  &  native & 49.9 & 48.7 & \textbf{50.1} \\
Gemma-7B &  cohere & 49.7 & 48.3 & \ul{50.4} \\
  &  google & 49.8 & 46.4 & \ul{50.7} \\
\cdashlinelr{1-5}
  &  native & 72.0 & \textbf{72.3} & \textbf{72.6} \\
Qwen1.5-7B &  cohere & 71.9 & 68.4 & 71.4 \\
  &  google & 71.9 & 67.6 & 71.4 \\
\cdashlinelr{1-5}
  &  native & 77.7 & \textbf{78.2} & \textbf{78.2} \\
Qwen1.5-14B &  cohere & 77.8 & 76.2 & 77.6 \\
  &  google & 77.8 & 75.8 & 77.2 \\
\bottomrule
\end{tabular}
\vspace{-1ex}
\caption{Sometimes the gap can be closed on CMMLU.}
\end{table}

\paragraph{Setup} For each of our previous data-model combinations, we now have two variants. Due to the space constraint, we only display results from larger models in the main text for the following benchmarks: TyDi QA, CMMLU, XQuAD, MSGM, MT-MMLU, and open-ended question answering. We \textbf{bold} the best native results and \ul{underline} translated results if they are close to native---meaning that the gap can be closed. Moreover, exhaustive results for all models and all languages on all benchmarks are enclosed in Tables~\ref{tab:tydiqa-russian-full}~to~\ref{tab:qa_native_cmdr+} in \Cref{app:full-results}.

\begin{table}[t]
\setlength{\tabcolsep}{4.5pt}
\centering\small
\begin{tabular}{llccc}
\toprule
\multicolumn{1}{c}{\multirow[c]{2}{*}{Base Model}}  & \multicolumn{1}{c}{\multirow[c]{2}{*}{Data}} & $10^{-6}$$\leftarrow$ & $10^{-4}$ & $10^{-4}$ \\
  &  & Mono & Mono & $\rightarrow$Multi\\
\midrule
            &  native & 18.5 & \textbf{30.3} & \textbf{31.0} \\
Llama2-7B &  cohere & 18.0 & 20.8 & 21.6 \\
            &  google & 17.8 & 21.1 & 24.1 \\
\cdashlinelr{1-5}
            &  native & \textbf{17.8} & \textbf{17.4} & 16.8 \\
Gemma-7B &  cohere & 17.3 & 14.8 & 16.3 \\
            &  google & 17.2 & 14.5 & 15.3 \\
\cdashlinelr{1-5}
                &  native & 30.7 & \textbf{34.9} & \textbf{42.6} \\
   Qwen1.5-7B &  cohere & 30.2 & 24.9 & 31.5 \\
                &  google & 29.9 & 22.0 & 27.7 \\
\cdashlinelr{1-5}
                &  native & 33.4 & \textbf{36.5} & \textbf{45.6} \\
  Qwen1.5-14B &  cohere & 33.6 & 28.5 & 30.8 \\
                &  google & 33.5 & 26.4 & 32.2 \\
\bottomrule
\end{tabular}
\vspace{-1ex}
\caption{There is always a large gap on XQuAD (EM).}
\vspace{-1ex}
\label{tab:xquad-small}
\end{table}

\begin{table}[t]
\setlength{\tabcolsep}{4.5pt}
\centering\small
\begin{tabular}{llccc}
\toprule
\multicolumn{1}{c}{\multirow[c]{2}{*}{Base Model}}  & \multicolumn{1}{c}{\multirow[c]{2}{*}{Data}} & $10^{-6}$$\leftarrow$ & $10^{-4}$ & $10^{-4}$ \\
  &  & Mono & Mono & $\rightarrow$Multi\\
\midrule
  &  native & \phantom{0}9.5 & \phantom{0}9.3 & 10.8\\
Llama2-7B &  cohere & \phantom{0}9.8 & \phantom{0}9.1 & 10.8 \\
  &  google & \phantom{0}9.7 & \phantom{0}8.1 & 12.0 \\
\cdashlinelr{1-5}
  &  native & 38.9 & 32.5 & 37.1 \\
Gemma-7B &  cohere & 38.8 & 33.6 & 37.2 \\
  &  google & 39.5 & 36.8 & 36.4 \\
\cdashlinelr{1-5}
  &  native & 42.1 & 40.9 & 41.2 \\
Qwen1.5-7B &  cohere & 41.5 & 38.1 & 40.8 \\
  &  google & 43.3 & 40.8 & 44.5 \\
\cdashlinelr{1-5}
  &  native & 55.9 & 52.8 & 55.2 \\
Qwen1.5-14B &  cohere & 55.7 & 49.1 & 53.5 \\
  &  google & 55.7 & 52.1 & 56.4 \\
\bottomrule
\end{tabular}
\vspace{-1ex}
\caption{There is always no gap on MGSM}
\vspace{-2ex}
\label{tab:mgsm-small}
\end{table}

\begin{table}[t]
\setlength{\tabcolsep}{4.5pt}
\centering\small
\begin{tabular}{llccc}
\toprule
\multicolumn{1}{c}{\multirow[c]{2}{*}{Base Model}}  & \multicolumn{1}{c}{\multirow[c]{2}{*}{Data}} & $10^{-6}$$\leftarrow$ & $10^{-4}$ & $10^{-4}$ \\
  &  & Mono & Mono & $\rightarrow$Multi\\
\midrule
  &  native & 35.8 & 35.6 & 36.3 \\
Llama2-7B &  cohere & 35.8 & 33.4 & 34.1 \\
  &  google & 35.8 & 33.7 & 34.3 \\
\cdashlinelr{1-5}
  &  native & 53.7 & 52.5 & 53.7 \\
Gemma-7B &  cohere & 53.8 & 54.4 & 55.6 \\
  &  google & 54.0 & 53.1 & 55.6 \\
\cdashlinelr{1-5}
  &  native & 50.3 & 50.2 & 50.6 \\
Qwen1.5-7B &  cohere & 50.2 & 49.6 & 51.2 \\
  &  google & 50.1 & 49.8 & 50.9 \\
\cdashlinelr{1-5}
  &  native & 58.2 & 58.3 & 58.3 \\
Qwen1.5-14B &  cohere & 58.3 & 56.3 & 58.5 \\
  &  google & 58.3 & 56.1 & 58.2 \\
\bottomrule
\end{tabular}
\vspace{-1ex}
\caption{There is always no gap on MT-MMLU.}
\vspace{-1ex}
\label{tab:mmmlu-small}
\end{table}

\paragraph{Native, structured benchmarks} We make bold those scores that are higher than the rest for each model under all hyperparameter settings. We find that the pattern seems to be affected by the base model and the task. It can be seen that Llama2-7B and Gemma-7B enjoy a performance leap in two scenarios: 1) on TyDi QA with a lower learning rate; and 2) on CMMLU with multilingual instruction tuning. In both cases, the performance gap between native and translated data can be overcome. However, for Qwen1.5, while the results change as the training configuration changes, native data still is the best data condition to go with.

\paragraph{Translated, structured benchmarks} Moving on to the translated test set results listed in \Cref{tab:xquad-small,tab:mgsm-small,tab:mmmlu-small}, we find that our previous findings still apply even when the learning rate is lowered or multilingual instruction tuning is applied. It can be seen that for the generative XQuAD, most of the time native instruction data maintains a huge advantage over the other two translated versions. Nonetheless, for MGSM and MT-MMLU, the difference between using translated and native data is not clear under most conditions. These also indicate that the stability of our results on translated structured tasks is not affected by the two hyperparameters.

\begin{table}[t]
\setlength{\tabcolsep}{8pt}
\centering\small
\begin{tabular}{llcc}
\toprule
\multicolumn{1}{c}{Base Model}  & \multicolumn{1}{c}{Data} & Mono & Multi\\
\midrule
           &  native & \textbf{171.5} & 121.7 \\
Llama2-7B  &  cohere & 126.3 & 126.3 \\
           &  google & 125.7 & 131.0 \\
\cdashlinelr{1-4}
           &  native & \textbf{216.5} & 164.7 \\
Gemma-7B   &  cohere & 150.0 & 146.3 \\
           &  google & 157.3 & 147.3 \\
\cdashlinelr{1-4}
            &  native & \textbf{187.5} & \textbf{189.3} \\
Qwen1.5-7B  &  cohere & 137.2 & 138.0 \\
            &  google & 132.7 & 133.7 \\
\cdashlinelr{1-4}
            &  native & \textbf{204.2} & \textbf{210.2} \\
Qwen1.5-14B &  cohere & 152.7 & 145.7 \\
            &  google & 140.0 & 140.7\\
\bottomrule
\end{tabular}
\vspace{-1ex}
\caption{There is always a large gap on open-ended question answering (translated, GPT-4-Turbo judged).}
\vspace{-2ex}
\label{tab:open-ended-translated-qa-small}
\end{table}

\paragraph{Open-ended question answering with translated questions} Finally, we compare monolingual and multilingual training on open-ended generation in \Cref{tab:open-ended-translated-qa-small}. Despite some fluctuations, the native-translated gap cannot be mitigated when evaluated on open-ended generation with translated questions. This is consistent with patterns on XQuAD that generative benchmarks can more effectively differentiate the instruction tuning data source.

\section{Conclusion and Future Work}

This work systematically analysed the effects of native and translated data on both the LLM instruction tuning and evaluation ends. The difference in data leads to result gaps on native test sets and generative benchmarks. We showed that knowledge mismatch is more likely to cause performance degradation rather than translation errors. With regularization, translated instruction data can potentially catch up with native data on structured benchmarks but not generative tasks.

Given our findings, we would like to call for prudent choices in multilingual LLM benchmarking. While the current work provides comprehensive empirical results and extrinsic evaluation, future work can consider investigating the knowledge in data intrinsically. More broadly, it is meaningful to coordinate efforts to develop large-scale native test sets that more accurately assess the breadth of languages and cultures LLMs aim to serve.

\section*{Limitations}
This paper focused on providing empirical results as an extrinsic evaluation of data characteristics. It can benefit from having an intrinsic understanding of the distinction between native and translated data, e.g. the knowledge or language features missing in the translated data and how this is associated with errors in specific test questions.

Also, our work centred around instruction tuning, but we have very limited knowledge of the pre-training data for the LLMs we study. This work assumes that the base models are described accurately by respective makers and that the LLM pre-training data would not prevent us from making meaningful scientific conclusions.

\section*{Ethical Considerations}
We consider our work to have minimal ethical risks. Like most papers on LLMs, it is difficult to make sure that the fine-tuned model is safe in all cases, but our models are not intended for the public. In terms of LLM evaluation, we believe this paper makes a positive contribution towards trustworthy and tailored evaluation for languages covered in large language models.

\section*{Acknowledgments}
The work has received funding from UK Research and Innovation (UKRI) under the UK government’s Horizon Europe funding guarantee [grant number 10052546].

Computations described in this research were supported by the Baskerville Tier 2 HPC service (https://www.baskerville.ac.uk/). Baskerville was funded by the EPSRC and UKRI through the World Class Labs scheme (EP/T022221/1) and the Digital Research Infrastructure programme (EP/W032244/1) and is operated by Advanced Research Computing at the University of Birmingham. We also acknowledge the Edinburgh International Data Facility (EIDF) and the Data-Driven Innovation Programme at the University of Edinburgh. The Cohere API credits were supported by a Cohere For AI Research Grant.

\bibliography{custom}

\appendix

\section{Prompts}
\subsection{Command R translation prompt}
\label{app:translation-prompt}

We list the translation prompt we use to query Command R in \Cref{fig:command-r-translation}, which asks the LLM to translate a given text while preserving the formatting. The source language, target language, and text variables are replaced by their string values during prompting.

\begin{figure}[htb]
  \centering\small\small
  \noindent\framebox{%
  \parbox{0.46\textwidth}{
  \texttt{Please translate from \$\{source\_lang\} to \$\{target\_lang\}. Do your best to preserve the formatting. The following content should and should only be translated.}\\
  
  \texttt{\$\{text\}}
  }%
  }
\caption{Prompt template for requesting a translation from Command R.}
\label{fig:command-r-translation}
\end{figure}

\subsection{LLM-as-a-judge prompt}
\label{app:llm-as-a-judge-prompt}

We list the LLM-as-a-judge prompt we use to query GPT-4-Turbo and Command-R+ in \Cref{fig:llm-eval-instruction}, which requires the judge to give a brief explanation before scoring. The instruction and response variables are replaced by their string values during prompting.

\begin{figure}[htb]
  \centering\small\small
  \noindent\framebox{%
  \parbox{0.46\textwidth}{
  \texttt{Please act as an impartial judge and evaluate the quality of a response to a user instruction displayed below. Your evaluation should consider factors such as helpfulness, relevance, accuracy, depth, creativity, and level of detail. Begin your evaluation with a brief explanation. After that, please rate the response on a scale of 1 to 5 by strictly following this format: ``[[rating]]''. The rating must be enclosed by two square brackets, for example: ``Rating: [[2]]''.}\\
  
  \texttt{[User Instruction]}
  
  \texttt{\$\{instruction\}}
  \\
  
  \texttt{[Response]}
  
  \texttt{\$\{response\}}
  }%
  }
\caption{Prompt template for requesting a model response evaluation from GPT-4-Turbo or Command-R+.}
\label{fig:llm-eval-instruction}
\end{figure}

\newpage
\section{Comprehensive Results}\label{app:comprehensive_results}

We list a breakdown of the results for each model and each language on various benchmarks in this appendix section. These are:
\begin{itemize}
    \item \Cref{tab:tydiqa-russian-full}: TyDi QA Russian, F1
    \item \Cref{tab:cmmlu-full}: CMMLU, accuracy
    \item \Cref{tab:xquad-1e-4-exact-match-full}: XQuAD, $10^{-4}$, exact match
    \item \Cref{tab:xquad-1e-6-exact-match-full}: XQuAD, $10^{-6}$, exact match
    \item \Cref{tab:xquad-1e-4-include-full}: XQuAD, $10^{-4}$, ``include''
    \item \Cref{tab:xquad-1e-6-include-full}: XQuAD, $10^{-6}$, ``include''
    \item \Cref{tab:mgsm-1e-4-full}: MGSM, $10^{-4}$, exact token match
    \item \Cref{tab:mgsm-1e-6-full}: MGSM, $10^{-6}$, exact token match
    \item \Cref{tab:mmmlu-1e-4}: MT-MMLU, $10^{-4}$, accuracy
    \item \Cref{tab:mmmlu-1e-6}: MT-MMLU, $10^{-6}$, accuracy
    \item \Cref{tab:openai_mmlu-1e-4}: HT-MMLU, $10^{-4}$, accuracy
    \item \Cref{tab:openai_mmlu-1e-6}: HT-MMLU, $10^{-6}$, accuracy
    \item \Cref{tab:qa_translated_gpt4}: translated questions, GPT-4 judge
    \item \Cref{tab:qa_translated_cmdr+}: translated questions, Cmd R+ judge
    \item \Cref{tab:qa_native_gpt4}: native questions, GPT-4 judge
    \item \Cref{tab:qa_native_cmdr+}: native questions, Cmd R+ judge
\end{itemize}

\label{app:full-results}
\begin{table*}[t]
\centering\small

\begin{tabular}{llccccc}
\toprule
\multirow{2}{*}{\text{Base Model}} & \multirow{2}{*}{\text{Data}} & \multicolumn{3}{c}{$1e{-4}$} & \multicolumn{2}{c}{$1e{-6}$} \\
\cmidrule(lr){3-5} \cmidrule(lr){6-7}
& & \text{Mono} & \text{Multi} & \text{RTT} & \text{Mono} & \text{Multi} \\
\midrule
& \text{native} & 28.3 & 25.1 & -- & 33.4 & 33.4 \\
\text{Llama2-7B} & \text{cohere} & 28.8 & 23.4 & 25.9 & 33.3 & 33.4 \\
& \text{google} & 25.5 & 22.9 & 25.7 & 33.3 & 33.4 \\
\cdashlinelr{1-7}
& \text{native} & 28.5 & 24.9 & -- & 27.7 & 27.6 \\
\text{Gemma-2B} & \text{cohere} & 28.2 & 28.8 & 27.7 & 27.7 & 27.8 \\
& \text{google} & 28.3 & 28.8 & 28.1 & 28.0 & 27.5 \\
\cdashlinelr{1-7}
& \text{native} & 33.6 & 31.5 & -- & 37.7 & 38.1 \\
\text{Gemma-7B} & \text{cohere} & 34.4 & 31.4 & 29.4 & 38.1 & 39.0 \\
& \text{google} & 30.7 & 30.9 & 33.3 & 37.9 & 38.7 \\
\cdashlinelr{1-7}
& \text{native} & 22.5 & 23.9 & -- & 16.6 & 17.1 \\
\text{Qwen1.5-0.5B} & \text{cohere} & 19.8 & 20.0 & 23.9 & 16.9 & 17.7 \\
& \text{google} & 17.5 & 19.8 & 23.0 & 16.8 & 17.1 \\
\cdashlinelr{1-7}
& \text{native} & 20.1 & 27.6 & -- & 19.5 & 19.7 \\
\text{Qwen1.5-1.8B} & \text{cohere} & 20.2 & 18.1 & 26.4 & 19.7 & 19.6 \\
& \text{google} & 18.1 & 19.3 & 23.6 & 19.8 & 19.8 \\
\cdashlinelr{1-7}
& \text{native} & 21.9 & 28.3 & -- & 18.1 & 18.2 \\
\text{Qwen1.5-4B} & \text{cohere} & 20.0 & 22.5 & 23.0 & 17.9 & 18.2 \\
& \text{google} & 20.1 & 22.5 & 22.4 & 17.9 & 18.2 \\
\cdashlinelr{1-7}
& \text{native} & 37.0 & 37.0 & -- & 22.4 & 23.8 \\
\text{Qwen1.5-7B} & \text{cohere} & 34.2 & 33.0 & 35.5 & 22.9 & 23.9 \\
& \text{google} & 27.2 & 27.1 & 34.9 & 22.7 & 23.9 \\
\cdashlinelr{1-7}
& \text{native} & 34.4 & 32.8 & -- & 24.8 & 25.1 \\
\text{Qwen1.5-14B} & \text{cohere} & 33.0 & 29.3 & 30.4 & 24.6 & 25.1 \\
& \text{google} & 32.1 & 35.2 & 30.7 & 24.9 & 25.9 \\
\bottomrule
\end{tabular}

\caption{Results for each model on TyDiQA Russian (F1, \%).}
\label{tab:tydiqa-russian-full}
\end{table*}

\begin{table*}[t]
\centering\small

\begin{tabular}{llccccc}
\toprule
\multirow{2}{*}{\text{Base Model}} & \multirow{2}{*}{\text{Data}} & \multicolumn{3}{c}{$1e{-4}$} & \multicolumn{2}{c}{$1e{-6}$} \\
\cmidrule(lr){3-5} \cmidrule(lr){6-7}
& & \text{Mono} & \text{Multi} & \text{RTT} & \text{Mono} & \text{Multi} \\
\midrule
& \text{native} & 32.7 & 32.6 & - & 31.8 & 31.9 \\
\text{Llama2-7B} & \text{cohere} & 30.2 & 32.7 & 31.6 & 32.0 & 31.6 \\
& \text{google} & 31.2 & 32.1 & 32.2 & 32.0 & 31.8 \\
\cdashlinelr{1-7}
& \text{native} & 31.8 & 31.5 & - & 31.2 & 31.0 \\
\text{Gemma-2B} & \text{cohere} & 29.4 & 31.2 & 30.4 & 31.2 & 31.0 \\
& \text{google} & 30.7 & 31.8 & 30.2 & 31.3 & 31.2 \\
\cdashlinelr{1-7}
& \text{native} & 48.7 & 50.1 & - & 49.9 & 49.7 \\
\text{Gemma-7B} & \text{cohere} & 48.3 & 50.4 & 48.6 & 49.7 & 49.9 \\
& \text{google} & 46.4 & 50.7 & 48.4 & 49.8 & 50.1 \\
\cdashlinelr{1-7}
& \text{native} & 44.1 & 43.9 & - & 42.2 & 42.1 \\
\text{Qwen1.5-0.5B} & \text{cohere} & 41.6 & 42.3 & 41.0 & 42.1 & 42.2 \\
& \text{google} & 42.8 & 42.7 & 43.0 & 42.0 & 42.2 \\
\cdashlinelr{1-7}
& \text{native} & 55.9 & 56.6 & - & 56.7 & 56.7 \\
\text{Qwen1.5-1.8B} & \text{cohere} & 53.8 & 56.6 & 54.7 & 56.6 & 56.4 \\
& \text{google} & 52.9 & 57.0 & 55.3 & 56.9 & 56.6 \\
\cdashlinelr{1-7}
& \text{native} & 65.3 & 66.4 & - & 66.0 & 65.8 \\
\text{Qwen1.5-4B} & \text{cohere} & 59.3 & 65.2 & 63.7 & 65.7 & 65.7 \\
& \text{google} & 59.8 & 65.2 & 64.6 & 66.0 & 65.8 \\
\cdashlinelr{1-7}
& \text{native} & 72.3 & 72.6 & - & 72.0 & 71.9 \\
\text{Qwen1.5-7B} & \text{cohere} & 68.4 & 71.4 & 68.9 & 71.9 & 71.8 \\
& \text{google} & 67.6 & 71.4 & 70.5 & 71.9 & 72.0 \\
\cdashlinelr{1-7}
& \text{native} & 78.2 & 78.2 & - & 77.7 & 77.6 \\
\text{Qwen1.5-14B} & \text{cohere} & 76.2 & 77.6 & 77.5 & 77.8 & 77.7 \\
& \text{google} & 75.8 & 77.2 & 77.4 & 77.8 & 77.7 \\
\bottomrule
\end{tabular}
\caption{Results for each model on CMMLU (accuracy, \%).}
\label{tab:cmmlu-full}
\end{table*}

\begin{table*}[htb]
\centering\small

\begin{tabular}{llcccGcccG}
\toprule
\multicolumn{1}{c}{\multirow[b]{2}{*}{Base model}} & \multicolumn{1}{c}{\multirow[b]{2}{*}{Data}} & \multicolumn{4}{c}{Monolingual} & \multicolumn{4}{c}{Multilingual} \\
\cmidrule(lr){3-6}\cmidrule(lr){7-10}
 & & es & ru & zh & \textit{average} & es & ru & zh & \textit{average} \\
\midrule
\multirow{3}{*}{Llama2-7B} 
 & native & 30.7 & 18.7 & 41.3 & 30.3 & 28.0 & 21.8 & 43.0 & 31.0 \\
 & cohere & 17.2 & 15.3 & 29.8 & 20.8 & 18.4 & 7.6 & 38.8 & 21.6 \\
 & google & 20.8 & 14.3 & 28.2 & 21.1 & 22.9 & 10.3 & 39.2 & 24.1 \\
\cdashlinelr{1-10}
\multirow{3}{*}{Gemma-2B}
 & native & 11.3 & 5.9 & 15.0 & 10.7 & 11.1 & 6.0 & 14.1 & 10.4 \\
 & cohere & 10.8 & 6.0 & 11.6 & 9.5 & 11.2 & 5.0 & 9.2 & 8.5 \\
 & google & 10.4 & 5.8 & 13.4 & 9.9 & 9.9 & 4.8 & 5.4 & 6.7 \\
\cdashlinelr{1-10}
\multirow{3}{*}{Gemma-7B}
 & native & 12.4 & 10.3 & 29.4 & 17.4 & 13.0 & 10.8 & 26.7 & 16.8 \\
 & cohere & 12.7 & 10.1 & 21.5 & 14.8 & 12.9 & 9.0 & 27.1 & 16.3 \\
 & google & 12.4 & 8.8 & 22.3 & 14.5 & 13.2 & 8.5 & 24.1 & 15.3 \\
\cdashlinelr{1-10}
\multirow{3}{*}{Qwen1.5-0.5B}
 & native & 26.5 & 7.6 & 20.6 & 18.2 & 18.1 & 10.2 & 21.7 & 16.6 \\
 & cohere & 9.8 & 6.6 & 16.1 & 10.8 & 11.3 & 7.4 & 15.5 & 11.4 \\
 & google & 10.8 & 6.5 & 17.2 & 11.5 & 12.7 & 7.7 & 16.1 & 12.2 \\
\cdashlinelr{1-10}
\multirow{3}{*}{Qwen1.5-1.8B}
 & native & 28.7 & 10.9 & 23.3 & 21.0 & 28.7 & 20.7 & 31.9 & 27.1 \\
 & cohere & 11.7 & 5.9 & 14.7 & 10.8 & 14.3 & 5.5 & 21.9 & 13.9 \\
 & google & 10.3 & 5.8 & 11.6 & 9.2 & 15.9 & 6.3 & 25.5 & 15.9 \\
\cdashlinelr{1-10}
\multirow{3}{*}{Qwen1.5-4B}
 & native & 33.1 & 24.9 & 37.7 & 31.9 & 36.5 & 31.4 & 52.6 & 40.2 \\
 & cohere & 29.6 & 19.3 & 25.4 & 24.8 & 31.9 & 18.4 & 39.4 & 29.9 \\
 & google & 19.7 & 18.7 & 23.9 & 20.8 & 35.1 & 20.3 & 40.2 & 31.8 \\
\cdashlinelr{1-10}
\multirow{3}{*}{Qwen1.5-7B}
 & native & 43.9 & 30.4 & 30.3 & 34.9 & 39.9 & 35.4 & 52.5 & 42.6 \\
 & cohere & 27.6 & 25.5 & 21.6 & 24.9 & 36.3 & 22.9 & 35.4 & 31.5 \\
 & google & 23.6 & 21.0 & 21.5 & 22.0 & 32.1 & 19.9 & 31.0 & 27.7 \\
\cdashlinelr{1-10}
\multirow{3}{*}{Qwen1.5-14B}
 & native & 44.7 & 34.2 & 30.7 & 36.5 & 49.7 & 38.7 & 48.3 & 45.6 \\
 & cohere & 35.3 & 28.4 & 21.9 & 28.5 & 39.9 & 24.0 & 28.6 & 30.8 \\
 & google & 28.3 & 26.8 & 24.2 & 26.4 & 42.1 & 26.6 & 27.7 & 32.2 \\
\bottomrule
\end{tabular}

\caption{All model and all language results on XQuAD ($10^{-4}$, exact match, \%).}
\label{tab:xquad-1e-4-exact-match-full}
\end{table*}

\begin{table*}[htb]
\centering\small

\begin{tabular}{llcccGcccG}
\toprule
\multicolumn{1}{c}{\multirow[b]{2}{*}{Base model}} & \multicolumn{1}{c}{\multirow[b]{2}{*}{Data}} & \multicolumn{4}{c}{Monolingual} & \multicolumn{4}{c}{Multilingual} \\
\cmidrule(lr){3-6}\cmidrule(lr){7-10}
 & & es & ru & zh & \textit{average} & es & ru & zh & \textit{average} \\
\midrule
\multirow{3}{*}{Llama2-7B} 
 & native & 12.9 & 7.3 & 35.3 & 18.5 & 12.4 & 7.1 & 31.9 & 17.1 \\
 & cohere & 12.9 & 7.1 & 33.9 & 18.0 & 12.4 & 7.3 & 32.3 & 17.3 \\
 & google & 12.9 & 7.1 & 33.3 & 17.8 & 12.4 & 7.2 & 32.2 & 17.3 \\
\cdashlinelr{1-10}
\multirow{3}{*}{Gemma-2B}
 & native & 10.0 & 6.1 & 4.8 & 7.0 & 10.0 & 6.2 & 4.9 & 7.0 \\
 & cohere & 10.0 & 5.9 & 4.6 & 6.8 & 9.9 & 6.1 & 4.8 & 6.9 \\
 & google & 10.0 & 6.1 & 4.7 & 6.9 & 10.0 & 6.0 & 4.9 & 6.9 \\
\cdashlinelr{1-10}
\multirow{3}{*}{Gemma-7B}
 & native & 12.5 & 9.1 & 31.7 & 17.8 & 12.5 & 9.1 & 31.2 & 17.6 \\
 & cohere & 12.4 & 9.1 & 30.6 & 17.3 & 12.4 & 9.3 & 30.3 & 17.3 \\
 & google & 12.3 & 9.1 & 30.3 & 17.2 & 12.9 & 9.1 & 30.3 & 17.4 \\
\cdashlinelr{1-10}
\multirow{3}{*}{Qwen1.5-0.5B}
& native & 12.1 & 6.2 & 12.8 & 10.4 & 10.5 & 6.5 & 11.2 & 9.4 \\
 & cohere & 10.8 & 6.2 & 11.8 & 9.6 & 10.2 & 6.5 & 10.6 & 9.1 \\
 & google & 11.1 & 6.2 & 12.0 & 9.8 & 9.9 & 6.2 & 11.0 & 9.0 \\
\cdashlinelr{1-10}
\multirow{3}{*}{Qwen1.5-1.8B}
 & native & 15.6 & 5.1 & 20.4 & 13.7 & 14.1 & 5.1 & 16.0 & 11.7 \\
 & cohere & 15.0 & 5.2 & 18.7 & 13.0 & 14.1 & 5.0 & 16.3 & 11.8 \\
 & google & 14.9 & 5.0 & 18.2 & 12.7 & 14.0 & 5.0 & 16.5 & 11.8 \\
\cdashlinelr{1-10}
\multirow{3}{*}{Qwen1.5-4B}
 & native & 24.4 & 14.3 & 44.9 & 27.8 & 21.0 & 14.9 & 37.8 & 24.6 \\
 & cohere & 23.3 & 14.2 & 43.4 & 26.9 & 20.9 & 14.7 & 37.6 & 24.4 \\
 & google & 22.7 & 14.3 & 43.5 & 26.8 & 20.8 & 15.0 & 37.6 & 24.5 \\
\cdashlinelr{1-10}
\multirow{3}{*}{Qwen1.5-7B}
 & native & 31.3 & 19.0 & 41.8 & 30.7 & 23.4 & 19.9 & 33.2 & 25.5 \\
 & cohere & 32.8 & 18.8 & 39.0 & 30.2 & 23.2 & 19.4 & 33.3 & 25.3 \\
 & google & 31.9 & 19.1 & 38.7 & 29.9 & 22.8 & 19.6 & 32.9 & 25.1 \\
\cdashlinelr{1-10}
\multirow{3}{*}{Qwen1.5-14B}
 & native & 37.7 & 26.9 & 35.7 & 33.4 & 37.2 & 27.6 & 28.1 & 31.0 \\
 & cohere & 38.3 & 27.1 & 35.3 & 33.6 & 36.7 & 27.5 & 27.3 & 30.5 \\
 & google & 37.6 & 27.0 & 35.9 & 33.5 & 36.8 & 27.3 & 27.4 & 30.5 \\
\bottomrule
\end{tabular}
\caption{Results for each model and each language on XQuAD ($10^{-6}$, exact match, \%).}
\label{tab:xquad-1e-6-exact-match-full}
\end{table*}

\begin{table*}[htb]
\centering\small

\begin{tabular}{llcccGcccG}
\toprule
\multicolumn{1}{c}{\multirow[b]{2}{*}{Base model}} & \multicolumn{1}{c}{\multirow[b]{2}{*}{Data}} & \multicolumn{4}{c}{Monolingual} & \multicolumn{4}{c}{Multilingual} \\
\cmidrule(lr){3-6}\cmidrule(lr){7-10}
 & & es & ru & zh & \textit{average} & es & ru & zh & \textit{average} \\
\midrule
\multirow{3}{*}{Llama2-7B} 
 & native & 51.7 & 31.8 & 66.0 & 49.8 & 55.6 & 30.8 & 61.9 & 49.4 \\
 & cohere & 23.5 & 26.3 & 43.4 & 31.1 & 30.3 & 16.9 & 55.3 & 34.2 \\
 & google & 43.7 & 27.1 & 43.8 & 38.2 & 46.1 & 28.4 & 58.7 & 44.4 \\
\cdashlinelr{1-10}
\multirow{3}{*}{Gemma-2B}
 & native & 17.2 & 18.8 & 50.3 & 28.8 & 19.6 & 13.9 & 44.7 & 26.1 \\
 & cohere & 26.3 & 20.3 & 36.6 & 27.7 & 36.1 & 17.8 & 47.1 & 33.7 \\
 & google & 33.7 & 22.6 & 39.7 & 32.0 & 39.2 & 22.8 & 52.8 & 38.3 \\
\cdashlinelr{1-10}
\multirow{3}{*}{Gemma-7B}
 & native & 14.7 & 24.7 & 61.8 & 33.7 & 15.7 & 17.5 & 67.2 & 33.5 \\
 & cohere & 13.9 & 20.3 & 37.5 & 23.9 & 13.4 & 12.3 & 44.6 & 23.4 \\
 & google & 16.5 & 29.0 & 48.5 & 31.3 & 23.7 & 19.3 & 48.2 & 30.4 \\
\cdashlinelr{1-10}
\multirow{3}{*}{Qwen1.5-0.5B}
 & native & 35.7 & 19.7 & 54.5 & 36.6 & 29.7 & 20.4 & 46.1 & 32.1 \\
 & cohere & 19.7 & 23.8 & 33.3 & 25.6 & 25.5 & 16.7 & 45.2 & 29.1 \\
 & google & 26.9 & 24.5 & 37.9 & 29.8 & 28.8 & 16.9 & 46.2 & 30.6 \\
\cdashlinelr{1-10}
\multirow{3}{*}{Qwen1.5-1.8B}
 & native & 44.9 & 25.8 & 64.4 & 45.0 & 45.0 & 30.6 & 56.4 & 44.0 \\
 & cohere & 24.0 & 20.8 & 40.6 & 28.5 & 34.2 & 20.0 & 56.2 & 36.8 \\
 & google & 35.5 & 19.5 & 46.8 & 33.9 & 35.4 & 23.2 & 59.1 & 39.2 \\
\cdashlinelr{1-10}
\multirow{3}{*}{Qwen1.5-4B}
 & native & 57.8 & 35.5 & 69.0 & 54.1 & 57.2 & 41.8 & 72.3 & 57.1 \\
 & cohere & 40.7 & 34.4 & 49.7 & 41.6 & 51.6 & 28.1 & 63.9 & 47.8 \\
 & google & 38.9 & 33.4 & 58.5 & 43.6 & 51.9 & 31.5 & 67.5 & 50.3 \\
\cdashlinelr{1-10}
\multirow{3}{*}{Qwen1.5-7B}
 & native & 58.6 & 44.0 & 68.4 & 57.0 & 65.0 & 45.5 & 72.6 & 61.0 \\
 & cohere & 48.9 & 33.5 & 44.6 & 42.4 & 43.5 & 31.7 & 61.6 & 45.6 \\
 & google & 44.2 & 37.4 & 54.7 & 45.4 & 44.9 & 33.9 & 62.5 & 47.1 \\
\cdashlinelr{1-10}
\multirow{3}{*}{Qwen1.5-14B}
 & native & 61.1 & 43.5 & 69.6 & 58.1 & 65.0 & 51.0 & 69.7 & 61.9 \\
 & cohere & 48.3 & 35.6 & 43.3 & 42.4 & 43.1 & 30.1 & 56.7 & 43.3 \\
 & google & 48.9 & 34.0 & 55.2 & 46.1 & 48.5 & 30.8 & 57.8 & 45.7 \\
\bottomrule
\end{tabular}
\caption{Results for each model and each language on XQuAD ($10^{-4}$, ``include'', \%).}
\label{tab:xquad-1e-4-include-full}
\end{table*}

\begin{table*}[htb]
\centering\small

\begin{tabular}{llcccGcccG}
\toprule
\multicolumn{1}{c}{\multirow[b]{2}{*}{Base model}} & \multicolumn{1}{c}{\multirow[b]{2}{*}{Data}} & \multicolumn{4}{c}{Monolingual} & \multicolumn{4}{c}{Multilingual} \\
\cmidrule(lr){3-6}\cmidrule(lr){7-10}
 & & es & ru & zh & \textit{average} & es & ru & zh & \textit{average} \\
\midrule
\multirow{3}{*}{Llama2-7B} 
 & native & 15.3 & 11.7 & 53.0 & 26.7 & 14.2 & 11.8 & 52.6 & 26.2 \\
 & cohere & 15.0 & 11.6 & 53.4 & 26.7 & 14.0 & 11.8 & 53.2 & 26.4 \\
 & google & 15.0 & 11.6 & 53.1 & 26.6 & 14.3 & 11.8 & 52.7 & 26.2 \\
\cdashlinelr{1-10}
\multirow{3}{*}{Gemma-2B}
 & native & 37.6 & 19.2 & 45.5 & 34.1 & 36.5 & 19.3 & 45.2 & 33.7 \\
 & cohere & 37.8 & 19.7 & 45.4 & 34.3 & 36.8 & 19.2 & 45.0 & 33.7 \\
 & google & 38.4 & 19.6 & 45.0 & 34.3 & 36.5 & 19.2 & 45.5 & 33.7 \\
\cdashlinelr{1-10}
\multirow{3}{*}{Gemma-7B}
 & native & 14.2 & 11.8 & 50.7 & 25.6 & 15.9 & 11.6 & 52.4 & 26.6 \\
 & cohere & 13.5 & 11.0 & 45.4 & 23.3 & 16.0 & 10.3 & 51.3 & 25.9 \\
 & google & 12.4 & 11.3 & 45.7 & 23.1 & 16.6 & 10.4 & 50.9 & 26.0 \\
\cdashlinelr{1-10}
\multirow{3}{*}{Qwen1.5-0.5B}
 & native & 41.1 & 31.1 & 53.9 & 42.0 & 43.9 & 30.8 & 55.4 & 43.4 \\
 & cohere & 43.2 & 32.2 & 54.9 & 43.4 & 44.5 & 31.9 & 55.7 & 44.1 \\
 & google & 42.8 & 31.8 & 55.5 & 43.3 & 44.1 & 31.6 & 55.7 & 43.8 \\
\cdashlinelr{1-10}
\multirow{3}{*}{Qwen1.5-1.8B}
 & native & 39.4 & 24.4 & 59.5 & 41.1 & 42.4 & 24.4 & 60.4 & 42.4 \\
 & cohere & 41.1 & 24.8 & 59.8 & 41.9 & 42.6 & 24.3 & 61.0 & 42.6 \\
 & google & 40.6 & 24.4 & 59.7 & 41.6 & 42.2 & 24.5 & 60.8 & 42.5 \\
\cdashlinelr{1-10}
\multirow{3}{*}{Qwen1.5-4B}
 & native & 59.4 & 37.7 & 70.5 & 55.9 & 62.1 & 37.3 & 71.0 & 56.8 \\
 & cohere & 59.7 & 37.2 & 70.7 & 55.9 & 62.4 & 37.4 & 71.3 & 57.0 \\
 & google & 60.3 & 36.6 & 70.6 & 55.8 & 62.4 & 37.8 & 70.8 & 57.0 \\
\cdashlinelr{1-10}
\multirow{3}{*}{Qwen1.5-7B}
 & native & 63.4 & 44.9 & 70.6 & 59.6 & 64.9 & 43.8 & 72.9 & 60.5 \\
 & cohere & 62.4 & 45.2 & 70.3 & 59.3 & 65.8 & 44.0 & 73.6 & 61.1 \\
 & google & 63.4 & 45.1 & 70.3 & 59.6 & 65.2 & 44.5 & 73.0 & 60.9 \\
\cdashlinelr{1-10}
\multirow{3}{*}{Qwen1.5-14B}
 & native & 57.1 & 43.4 & 70.8 & 57.1 & 59.7 & 43.3 & 72.6 & 58.5 \\
 & cohere & 56.1 & 43.8 & 70.3 & 56.7 & 59.8 & 43.4 & 72.9 & 58.7 \\
 & google & 56.6 & 43.4 & 70.2 & 56.8 & 59.0 & 43.1 & 72.4 & 58.2 \\
\bottomrule
\end{tabular}
\caption{Results for each model and each language on XQuAD ($10^{-6}$, ``include'', \%).}
\label{tab:xquad-1e-6-include-full}
\end{table*}

\begin{table*}[htb]
\centering\small

\begin{tabular}{llcccGcccG}
\toprule
\multicolumn{1}{c}{\multirow[b]{2}{*}{Base model}} & \multicolumn{1}{c}{\multirow[b]{2}{*}{Data}} & \multicolumn{4}{c}{Monolingual} & \multicolumn{4}{c}{Multilingual} \\
\cmidrule(lr){3-6}\cmidrule(lr){7-10}
 & & es & ru & zh & \textit{average} & es & ru & zh & \textit{average} \\
\midrule
\multirow{3}{*}{Llama2-7B} & native & 8.4 & 10.0 & 9.6 & 9.3 & 12.0 & 10.4 & 10.0 & 10.8 \\
 & cohere & 5.2 & 11.2 & 10.8 & 9.1 & 10.0 & 12.4 & 10.0 & 10.8 \\
 & google & 8.4 & 9.2 & 6.8 & 8.1 & 12.0 & 10.8 & 13.2 & 12.0 \\
\cdashlinelr{1-10}
\multirow{3}{*}{Gemma-2B} & native & 11.6 & 12.8 & 8.8 & 11.1 & 13.2 & 10.8 & 9.2 & 11.1 \\
 & cohere & 11.6 & 12.4 & 15.6 & 13.2 & 11.6 & 10.8 & 10.8 & 11.1 \\
 & google & 12.8 & 11.2 & 8.8 & 10.9 & 10.0 & 9.2 & 11.2 & 10.1 \\
\cdashlinelr{1-10}
\multirow{3}{*}{Gemma-7B} & native & 30.0 & 48.8 & 18.8 & 32.5 & 36.4 & 47.2 & 27.6 & 37.1 \\
 & cohere & 34.4 & 46.8 & 19.6 & 33.6 & 36.4 & 44.0 & 31.2 & 37.2 \\
 & google & 30.0 & 48.8 & 31.6 & 36.8 & 37.6 & 42.8 & 28.8 & 36.4 \\
\cdashlinelr{1-10}
\multirow{3}{*}{Qwen1.5-0.5B} & native & 2.8 & 2.0 & 4.8 & 3.2 & 2.0 & 1.6 & 10.4 & 4.7 \\
 & cohere & 1.6 & 2.4 & 8.0 & 4.0 & 3.6 & 2.4 & 8.8 & 4.9 \\
 & google & 2.0 & 2.0 & 9.2 & 4.4 & 2.4 & 3.6 & 7.6 & 4.5 \\
\cdashlinelr{1-10}
\multirow{3}{*}{Qwen1.5-1.8B} & native & 6.0 & 6.4 & 15.6 & 9.3 & 9.6 & 6.0 & 19.6 & 11.7 \\
 & cohere & 8.4 & 5.6 & 14.8 & 9.6 & 6.8 & 7.6 & 15.6 & 10.0 \\
 & google & 6.4 & 5.6 & 14.8 & 8.9 & 6.0 & 5.2 & 21.2 & 10.8 \\
\cdashlinelr{1-10}
\multirow{3}{*}{Qwen1.5-4B} & native & 18.0 & 24.0 & 34.4 & 25.5 & 22.0 & 26.0 & 40.4 & 29.5 \\
 & cohere & 20.0 & 21.6 & 16.0 & 19.2 & 21.6 & 24.8 & 42.0 & 29.5 \\
 & google & 17.6 & 22.4 & 35.6 & 25.2 & 21.2 & 21.6 & 38.0 & 26.9 \\
\cdashlinelr{1-10}
\multirow{3}{*}{Qwen1.5-7B}  & native & 40.4 & 40.8 & 41.6 & 40.9 & 34.8 & 40.0 & 48.8 & 41.2 \\
 & cohere & 37.2 & 40.4 & 36.8 & 38.1 & 37.2 & 39.6 & 45.6 & 40.8 \\
 & google & 42.4 & 40.8 & 39.2 & 40.8 & 42.8 & 42.4 & 48.4 & 44.5 \\
\cdashlinelr{1-10}
\multirow{3}{*}{Qwen1.5-14B} & native & 44.8 & 60.0 & 53.6 & 52.8 & 49.2 & 59.6 & 56.8 & 55.2 \\
 & cohere & 32.8 & 63.6 & 50.8 & 49.1 & 49.2 & 59.6 & 51.6 & 53.5 \\
 & google & 42.4 & 60.0 & 54.0 & 52.1 & 50.4 & 63.2 & 55.6 & 56.4 \\
\bottomrule
\end{tabular}
\caption{Results for each model and each language on MGSM ($10^{-4}$, exact token match, \%).}
\label{tab:mgsm-1e-4-full}
\end{table*}

\begin{table*}[htb]
\centering\small

\begin{tabular}{llcccGcccG}
\toprule
\multicolumn{1}{c}{\multirow[b]{2}{*}{Base model}} & \multicolumn{1}{c}{\multirow[b]{2}{*}{Data}} & \multicolumn{4}{c}{Monolingual} & \multicolumn{4}{c}{Multilingual} \\
\cmidrule(lr){3-6}\cmidrule(lr){7-10}
 & & es & ru & zh & \textit{average} & es & ru & zh & \textit{average} \\
\midrule
\multirow{3}{*}{Llama2-7B} & native & 9.6 & 10.0 & 8.8 & 9.5 & 10.0 & 9.6 & 8.8 & 9.5 \\
 & cohere & 10.4 & 11.2 & 8.0 & 9.9 & 10.0 & 10.4 & 9.2 & 9.9 \\
 & google & 9.6 & 9.6 & 10.0 & 9.7 & 10.4 & 9.6 & 9.2 & 9.7 \\
\cdashlinelr{1-10}
\multirow{3}{*}{Gemma-2B} & native & 13.2 & 10.8 & 10.4 & 11.5 & 12.4 & 11.2 & 12.4 & 12.0 \\
 & cohere & 12.4 & 10.4 & 12.0 & 11.6 & 12.4 & 11.6 & 12.8 & 12.3 \\
 & google & 13.6 & 11.6 & 13.2 & 12.8 & 14.4 & 11.6 & 11.6 & 12.5 \\
\cdashlinelr{1-10}
\multirow{3}{*}{Gemma-7B} & native & 34.8 & 44.8 & 37.2 & 38.9 & 36.8 & 46.4 & 36.4 & 39.9 \\
 & cohere & 35.6 & 44.4 & 36.4 & 38.8 & 34.8 & 45.2 & 38.0 & 39.3 \\
 & google & 35.2 & 46.8 & 36.4 & 39.5 & 37.2 & 44.0 & 36.8 & 39.3 \\
\cdashlinelr{1-10}
\multirow{3}{*}{Qwen1.5-0.5B} & native & 2.4 & 2.8 & 8.8 & 4.7 & 2.4 & 2.4 & 8.4 & 4.4 \\
 & cohere & 2.8 & 2.0 & 10.8 & 5.2 & 3.2 & 2.4 & 8.8 & 4.8 \\
 & google & 2.8 & 1.6 & 6.8 & 3.7 & 2.4 & 2.4 & 8.8 & 4.5 \\
\cdashlinelr{1-10}
\multirow{3}{*}{Qwen1.5-1.8B} & native & 7.2 & 6.0 & 20.8 & 11.3 & 7.2 & 6.8 & 24.4 & 12.8 \\
 & cohere & 8.0 & 6.8 & 20.4 & 11.7 & 5.6 & 6.0 & 23.2 & 11.6 \\
 & google & 6.0 & 5.6 & 20.8 & 10.8 & 6.8 & 6.0 & 24.4 & 12.4 \\
\cdashlinelr{1-10}
\multirow{3}{*}{Qwen1.5-4B} & native & 21.6 & 28.0 & 40.0 & 29.9 & 20.8 & 28.0 & 40.0 & 29.6 \\
 & cohere & 21.2 & 27.6 & 40.0 & 29.6 & 22.0 & 28.8 & 38.8 & 29.9 \\
 & google & 22.8 & 28.8 & 41.6 & 31.1 & 21.2 & 28.0 & 38.0 & 29.1 \\
\cdashlinelr{1-10}
\multirow{3}{*}{Qwen1.5-7B} & native & 38.0 & 41.6 & 46.8 & 42.1 & 35.2 & 42.4 & 53.2 & 43.6 \\
 & cohere & 36.0 & 41.2 & 47.2 & 41.5 & 34.8 & 40.0 & 48.8 & 41.2 \\
 & google & 38.4 & 43.2 & 48.4 & 43.3 & 34.0 & 43.6 & 50.8 & 42.8 \\
\cdashlinelr{1-10}
\multirow{3}{*}{Qwen1.5-14B} & native & 46.0 & 63.6 & 58.0 & 55.9 & 47.2 & 62.8 & 58.0 & 56.0 \\
 & cohere & 47.6 & 61.6 & 58.0 & 55.7 & 46.8 & 63.6 & 57.6 & 56.0 \\
 & google & 46.8 & 62.4 & 58.0 & 55.7 & 48.0 & 61.6 & 57.6 & 55.7 \\
\bottomrule
\end{tabular}
\caption{Results for each model and each language on MGSM ($10^{-6}$, exact token match, \%).}
\label{tab:mgsm-1e-6-full}
\end{table*}

\begin{table*}[htb]
\centering\small

\begin{tabular}{llcccGcccG}
\toprule
\multicolumn{1}{c}{\multirow[b]{2}{*}{Base model}} & \multicolumn{1}{c}{\multirow[b]{2}{*}{Data}} & \multicolumn{4}{c}{Monolingual} & \multicolumn{4}{c}{Multilingual} \\
\cmidrule(lr){3-6}\cmidrule(lr){7-10}
 & & es & ru & zh & \textit{average} & es & ru & zh & \textit{average} \\
\midrule
\multirow{3}{*}{Llama2-7B} 
 & native & 38.4 & 33.9 & 34.4 & 35.6 & 39.6 & 35.3 & 33.9 & 36.3 \\
 & cohere & 38.0 & 34.6 & 27.6 & 33.4 & 37.8 & 31.6 & 32.9 & 34.1 \\
 & google & 36.4 & 34.4 & 30.4 & 33.7 & 37.6 & 31.9 & 33.5 & 34.3 \\
\cdashlinelr{1-10}
\multirow{3}{*}{Gemma-2B}
 & native & 33.3 & 30.8 & 30.5 & 31.5 & 32.4 & 30.4 & 30.7 & 31.2 \\
 & cohere & 30.8 & 30.3 & 29.6 & 30.2 & 33.7 & 31.9 & 32.8 & 32.8 \\
 & google & 31.8 & 30.0 & 32.0 & 31.3 & 34.0 & 31.0 & 32.8 & 32.6 \\
\cdashlinelr{1-10}
\multirow{3}{*}{Gemma-7B}
 & native & 55.9 & 53.0 & 48.7 & 52.5 & 57.3 & 53.0 & 50.7 & 53.7 \\
 & cohere & 58.4 & 53.8 & 50.8 & 54.4 & 58.7 & 55.3 & 52.9 & 55.6 \\
 & google & 56.1 & 53.6 & 49.7 & 53.1 & 58.8 & 55.5 & 52.5 & 55.6 \\
\cdashlinelr{1-10}
\multirow{3}{*}{Qwen1.5-0.5B}
 & native & 30.2 & 27.1 & 35.3 & 30.9 & 29.4 & 26.5 & 35.2 & 30.4 \\
 & cohere & 30.1 & 26.5 & 35.5 & 30.7 & 28.7 & 26.8 & 34.8 & 30.1 \\
 & google & 32.4 & 26.2 & 36.9 & 31.8 & 29.5 & 26.6 & 34.6 & 30.2 \\
\cdashlinelr{1-10}
\multirow{3}{*}{Qwen1.5-1.8B}
 & native & 34.0 & 33.5 & 40.8 & 36.1 & 36.0 & 32.9 & 42.1 & 37.0 \\
 & cohere & 35.4 & 32.2 & 40.9 & 36.2 & 36.3 & 32.2 & 42.0 & 36.8 \\
 & google & 36.7 & 31.9 & 39.9 & 36.2 & 36.8 & 32.8 & 42.0 & 37.2 \\
\cdashlinelr{1-10}
\multirow{3}{*}{Qwen1.5-4B}
 & native & 40.9 & 38.2 & 49.3 & 42.8 & 45.5 & 38.7 & 49.6 & 44.6 \\
 & cohere & 39.9 & 39.4 & 44.5 & 41.3 & 43.9 & 37.8 & 49.4 & 43.7 \\
 & google & 39.6 & 38.9 & 44.2 & 40.9 & 43.3 & 36.2 & 49.0 & 42.8 \\
\cdashlinelr{1-10}
\multirow{3}{*}{Qwen1.5-7B}
 & native & 50.3 & 47.2 & 52.9 & 50.2 & 51.0 & 46.7 & 54.0 & 50.6 \\
 & cohere & 49.6 & 46.7 & 52.6 & 49.6 & 52.0 & 47.3 & 54.3 & 51.2 \\
 & google & 50.4 & 47.2 & 51.8 & 49.8 & 51.9 & 46.7 & 54.0 & 50.9 \\
\cdashlinelr{1-10}
\multirow{3}{*}{Qwen1.5-14B}
 & native & 58.1 & 55.4 & 61.3 & 58.3 & 58.6 & 54.9 & 61.5 & 58.3 \\
 & cohere & 55.8 & 55.2 & 57.9 & 56.3 & 59.3 & 55.3 & 61.1 & 58.5 \\
 & google & 54.9 & 55.5 & 58.0 & 56.1 & 59.1 & 55.2 & 60.4 & 58.2 \\
\bottomrule
\end{tabular}
\caption{Results for each model and each language on MT-MMLU ($10^{-4}$, accuracy, \%).}
\label{tab:mmmlu-1e-4}
\end{table*}

\begin{table*}[htb]
\centering\small

\begin{tabular}{llcccGcccG}
\toprule
\multicolumn{1}{c}{\multirow[b]{2}{*}{Base model}} & \multicolumn{1}{c}{\multirow[b]{2}{*}{Data}} & \multicolumn{4}{c}{Monolingual} & \multicolumn{4}{c}{Multilingual} \\
\cmidrule(lr){3-6}\cmidrule(lr){7-10}
 & & es & ru & zh & \textit{average} & es & ru & zh & \textit{average} \\
\midrule
\multirow{3}{*}{Llama2-7B} 
 & native & 39.1 & 34.5 & 33.9 & 35.8 & 38.9 & 34.7 & 33.7 & 35.8 \\
 & cohere & 39.1 & 34.5 & 33.7 & 35.8 & 39.0 & 34.5 & 33.5 & 35.7 \\
 & google & 39.1 & 34.5 & 33.9 & 35.8 & 38.9 & 34.3 & 33.6 & 35.6 \\
\cdashlinelr{1-10}
\multirow{3}{*}{Gemma-2B}
 & native & 31.8 & 30.8 & 31.7 & 31.4 & 31.7 & 30.6 & 31.6 & 31.3 \\
 & cohere & 31.8 & 30.8 & 31.7 & 31.4 & 32.0 & 30.6 & 31.6 & 31.4 \\
 & google & 32.0 & 30.8 & 31.7 & 31.5 & 31.8 & 30.7 & 31.7 & 31.4 \\
\cdashlinelr{1-10}
\multirow{3}{*}{Gemma-7B}
 & native & 56.5 & 53.6 & 51.0 & 53.7 & 56.7 & 53.7 & 51.5 & 54.0 \\
 & cohere & 56.5 & 53.6 & 51.2 & 53.8 & 56.8 & 54.1 & 51.3 & 54.1 \\
 & google & 56.2 & 53.9 & 51.7 & 54.0 & 57.3 & 54.0 & 51.2 & 54.1 \\
\cdashlinelr{1-10}
\multirow{3}{*}{Qwen1.5-0.5B}
 & native & 28.0 & 25.7 & 34.8 & 29.5 & 27.9 & 25.8 & 34.6 & 29.4 \\
 & cohere & 27.9 & 25.9 & 34.5 & 29.5 & 28.2 & 25.9 & 34.5 & 29.5 \\
 & google & 28.0 & 25.9 & 34.5 & 29.5 & 28.1 & 25.8 & 34.6 & 29.5 \\
\cdashlinelr{1-10}
\multirow{3}{*}{Qwen1.5-1.8B}
 & native & 36.1 & 31.8 & 41.4 & 36.4 & 35.9 & 31.8 & 41.6 & 36.5 \\
 & cohere & 36.0 & 31.7 & 41.5 & 36.4 & 36.0 & 31.7 & 41.5 & 36.4 \\
 & google & 36.2 & 31.8 & 41.3 & 36.4 & 36.1 & 31.7 & 41.3 & 36.4 \\
\cdashlinelr{1-10}
\multirow{3}{*}{Qwen1.5-4B}
 & native & 45.1 & 39.0 & 49.3 & 44.5 & 44.9 & 39.0 & 49.4 & 44.4 \\
 & cohere & 45.0 & 38.9 & 49.3 & 44.4 & 44.7 & 39.1 & 49.3 & 44.4 \\
 & google & 45.1 & 38.9 & 49.4 & 44.5 & 44.8 & 38.8 & 49.4 & 44.3 \\
\cdashlinelr{1-10}
\multirow{3}{*}{Qwen1.5-7B}
 & native & 51.3 & 46.4 & 53.3 & 50.3 & 51.0 & 46.3 & 53.1 & 50.1 \\
 & cohere & 51.2 & 46.4 & 53.1 & 50.2 & 51.1 & 46.4 & 53.2 & 50.2 \\
 & google & 51.2 & 46.3 & 52.9 & 50.1 & 51.0 & 46.2 & 53.2 & 50.2 \\
\cdashlinelr{1-10}
\multirow{3}{*}{Qwen1.5-14B}
 & native & 58.7 & 55.1 & 60.8 & 58.2 & 58.6 & 55.2 & 61.0 & 58.3 \\
 & cohere & 58.7 & 55.1 & 61.1 & 58.3 & 58.6 & 55.1 & 61.0 & 58.2 \\
 & google & 58.7 & 55.1 & 61.0 & 58.3 & 58.7 & 55.1 & 61.1 & 58.3 \\
\bottomrule
\end{tabular}
\caption{Results for each model and each language on MT-MMLU ($10^{-6}$, accuracy, \%).}
\label{tab:mmmlu-1e-6}
\end{table*}

\begin{table*}[htb]
\centering\small
\begin{tabular}{llccGccG}
\toprule
\multicolumn{1}{c}{\multirow[b]{2}{*}{Base Model}} & \multicolumn{1}{c}{\multirow[b]{2}{*}{Data}} & \multicolumn{3}{c}{Monolingual} & \multicolumn{3}{c}{Multilingual} \\
\cmidrule(lr){3-5}\cmidrule(lr){6-8}
 & & es & zh & \textit{average} & es & zh & \textit{average} \\
\midrule
\multirow{3}{*}{Llama2-7B} 
 & native & 37.6 & 33.8 & 35.7 & 39.0 & 33.5 & 36.3 \\
 & cohere & 37.2 & 27.8 & 32.5 & 36.9 & 32.1 & 34.5 \\
 & google & 35.9 & 29.5 & 32.7 & 37.1 & 32.5 & 34.8 \\
\cdashlinelr{1-8}
\multirow{3}{*}{Gemma-2B}
 & native & 33.3 & 31.1 & 32.2 & 32.3 & 31.0 & 31.7 \\
 & cohere & 30.2 & 29.4 & 29.8 & 33.3 & 32.4 & 32.9 \\
 & google & 31.1 & 31.8 & 31.4 & 33.6 & 33.0 & 33.3 \\
\cdashlinelr{1-8}
\multirow{3}{*}{Gemma-7B}
 & native & 54.9 & 48.0 & 51.5 & 56.2 & 50.4 & 53.3 \\
 & cohere & 57.5 & 50.3 & 53.9 & 57.8 & 53.1 & 55.4 \\
 & google & 55.6 & 48.8 & 52.2 & 57.7 & 52.7 & 55.2 \\
\cdashlinelr{1-8}
\multirow{3}{*}{Qwen1.5-0.5B}
 & native & 30.4 & 35.6 & 33.0 & 29.4 & 35.2 & 32.3 \\
 & cohere & 29.7 & 34.7 & 32.2 & 29.0 & 34.4 & 31.7 \\
 & google & 31.5 & 36.8 & 34.1 & 29.3 & 34.4 & 31.8 \\
\cdashlinelr{1-8}
\multirow{3}{*}{Qwen1.5-1.8B}
 & native & 33.2 & 40.7 & 37.0 & 35.8 & 42.6 & 39.2 \\
 & cohere & 35.2 & 40.7 & 38.0 & 36.7 & 42.0 & 39.3 \\
 & google & 36.3 & 40.0 & 38.1 & 37.1 & 42.3 & 39.7 \\
\cdashlinelr{1-8}
\multirow{3}{*}{Qwen1.5-4B}
 & native & 40.2 & 49.0 & 44.6 & 44.4 & 49.9 & 47.2 \\
 & cohere & 39.0 & 45.2 & 42.1 & 43.2 & 49.2 & 46.2 \\
 & google & 39.0 & 45.0 & 42.0 & 42.2 & 49.2 & 45.7 \\
\cdashlinelr{1-8}
\multirow{3}{*}{Qwen1.5-7B}
 & native & 49.6 & 53.0 & 51.3 & 50.4 & 53.4 & 51.9 \\
 & cohere & 48.4 & 51.8 & 50.1 & 50.5 & 54.3 & 52.4 \\
 & google & 49.3 & 51.3 & 50.3 & 50.5 & 53.3 & 51.9 \\
\cdashlinelr{1-8}
\multirow{3}{*}{Qwen1.5-14B}
 & native & 57.8 & 60.7 & 59.2 & 58.6 & 61.4 & 60.0 \\
 & cohere & 55.1 & 57.7 & 56.4 & 58.7 & 61.3 & 60.0 \\
 & google & 54.2 & 57.3 & 55.7 & 58.3 & 61.2 & 59.7 \\
\bottomrule
\end{tabular}

\caption{Results for each model and each language on HT-MMLU ($10^{-4}$, accuracy, \%).}
\label{tab:openai_mmlu-1e-4}
\end{table*}

\begin{table*}[htb]
\centering\small
\begin{tabular}{llccGccG}
\toprule
\multicolumn{1}{c}{\multirow[b]{2}{*}{Base Model}} & \multicolumn{1}{c}{\multirow[b]{2}{*}{Data}} & \multicolumn{3}{c}{Monolingual} & \multicolumn{3}{c}{Multilingual} \\
\cmidrule(lr){3-5}\cmidrule(lr){6-8}
 & & es & zh & \textit{average} & es & zh & \textit{average} \\
\midrule
\multirow{3}{*}{Llama-2-7B} 
 & native & 38.3 & 33.3 & 35.8 & 38.1 & 33.4 & 35.8 \\
 & cohere & 38.2 & 33.1 & 35.6 & 38.1 & 33.5 & 35.8 \\
 & google & 38.1 & 33.4 & 35.8 & 38.2 & 33.3 & 35.8 \\
\cdashlinelr{1-8}
\multirow{3}{*}{Gemma-2B}
 & native & 31.1 & 31.0 & 31.0 & 30.8 & 31.2 & 31.0 \\
 & cohere & 31.2 & 31.3 & 31.2 & 31.2 & 31.4 & 31.3 \\
 & google & 31.2 & 31.0 & 31.1 & 31.2 & 31.2 & 31.2 \\
\cdashlinelr{1-8}
\multirow{3}{*}{Gemma-7B}
 & native & 56.0 & 49.9 & 53.0 & 55.9 & 50.0 & 52.9 \\
 & cohere & 55.5 & 50.3 & 52.9 & 55.8 & 49.6 & 52.7 \\
 & google & 55.5 & 50.6 & 53.1 & 55.8 & 50.1 & 53.0 \\
\cdashlinelr{1-8}
\multirow{3}{*}{Qwen1.5-0.5B}
 & native & 28.5 & 34.6 & 31.5 & 28.4 & 34.6 & 31.5 \\
 & cohere & 28.5 & 34.7 & 31.6 & 28.4 & 34.7 & 31.5 \\
 & google & 28.4 & 34.5 & 31.4 & 28.4 & 34.6 & 31.5 \\
\cdashlinelr{1-8}
\multirow{3}{*}{Qwen1.5-1.8B}
 & native & 35.7 & 41.6 & 38.7 & 35.7 & 41.5 & 38.6 \\
 & cohere & 35.7 & 41.2 & 38.4 & 35.7 & 41.2 & 38.5 \\
 & google & 35.7 & 41.2 & 38.5 & 35.5 & 41.2 & 38.4 \\
\cdashlinelr{1-8}
\multirow{3}{*}{Qwen1.5-4B}
 & native & 43.7 & 49.6 & 46.7 & 43.9 & 49.5 & 46.7 \\
 & cohere & 43.9 & 49.4 & 46.6 & 43.7 & 49.4 & 46.5 \\
 & google & 43.8 & 49.6 & 46.7 & 43.7 & 49.4 & 46.5 \\
\cdashlinelr{1-8}
\multirow{3}{*}{Qwen1.5-7B}
 & native & 50.4 & 52.7 & 51.6 & 50.5 & 52.7 & 51.6 \\
 & cohere & 50.5 & 52.7 & 51.6 & 50.5 & 52.8 & 51.6 \\
 & google & 50.6 & 52.8 & 51.7 & 50.4 & 52.7 & 51.6 \\
\cdashlinelr{1-8}
\multirow{3}{*}{Qwen1.5-14B}
 & native & 58.3 & 61.0 & 59.6 & 58.5 & 61.1 & 59.8 \\
 & cohere & 58.5 & 61.2 & 59.9 & 58.6 & 61.3 & 59.9 \\
 & google & 58.4 & 61.1 & 59.8 & 58.5 & 61.2 & 59.9 \\
\bottomrule
\end{tabular}

\caption{Results for each model and each language on HT-MMLU ($10^{-6}$, accuracy, \%).}
\label{tab:openai_mmlu-1e-6}
\end{table*}

\begin{table*}[ht]
\centering\small
\begin{tabular}{llcccGcccG}
\toprule
\multicolumn{1}{c}{\multirow[b]{2}{*}{Base model}} & \multicolumn{1}{c}{\multirow[b]{2}{*}{Data}} & \multicolumn{4}{c}{Monolingual} & \multicolumn{4}{c}{Multilingual} \\
\cmidrule(lr){3-6}\cmidrule(lr){7-10}
 & & es & ru & zh & \textit{average} & es & ru & zh & \textit{average} \\
\midrule
 & native & 203.0 & 180.0 & 131.5 & 171.5 & 134.0 & 129.0 & 102.0 & 121.7 \\
Llama2-7B & cohere & 144.0 & 131.0 & 104.0 & 126.3 & 134.0 & 130.0 & 115.0 & 126.3 \\
 & google & 140.0 & 122.0 & 115.0 & 125.7 & 148.0 & 125.0 & 120.0 & 131.0 \\
\cdashlinelr{1-10}
 & native & 161.5 & 151.0 & 125.0 & 145.8 & 122.0 & 112.0 & 109.0 & 114.3 \\
Gemma-2B & cohere & 125.0 & 110.0 & 115.0 & 116.7 & 155.5 & 138.0 & 115.0 & 136.2 \\
 & google & 110.0 & 118.0 & 119.0 & 115.7 & 126.0 & 116.0 & 130.0 & 124.0 \\
\cdashlinelr{1-10}
 & native & 224.5 & 230.0 & 195.0 & 216.5 & 172.0 & 161.0 & 161.0 & 164.7 \\
Gemma-7B & cohere & 146.0 & 156.0 & 148.0 & 150.0 & 146.0 & 144.0 & 149.0 & 146.3 \\
 & google & 168.0 & 157.0 & 147.0 & 157.3 & 151.0 & 144.0 & 147.0 & 147.3 \\
\cdashlinelr{1-10}
 & native & \phantom{0}89.0 & \phantom{0}76.0 & 141.0 & 102.0 & \phantom{0}93.5 & \phantom{0}77.0 & 133.0 & 101.2 \\
Qwen1.5-0.5B & cohere & \phantom{0}70.0 & \phantom{0}60.0 & \phantom{0}99.0 & \phantom{0}76.3 & \phantom{0}78.0 & \phantom{0}61.0 & 107.0 & \phantom{0}82.0 \\
 & google & \phantom{0}75.0 & \phantom{0}61.0 & 109.0 & \phantom{0}81.7 & \phantom{0}72.0 & \phantom{0}58.0 & \phantom{0}88.0 & \phantom{0}72.7 \\
\cdashlinelr{1-10}
 & native & 119.5 & 112.0 & 148.0 & 126.5 & 105.0 & 104.0 & 162.0 & 123.7 \\
Qwen1.5-1.8B & cohere & \phantom{0}88.0 & \phantom{0}88.0 & 126.0 & 100.7 & \phantom{0}86.0 & \phantom{0}87.0 & 123.0 & \phantom{0}98.7 \\
 & google & \phantom{0}87.0 & \phantom{0}80.0 & 108.0 & \phantom{0}91.7 & 101.5 & \phantom{0}82.0 & 115.0 & \phantom{0}99.5 \\
\cdashlinelr{1-10}
 & native & 187.0 & 149.5 & 190.0 & 175.5 & 186.5 & 151.0 & 199.0 & 178.8 \\
Qwen1.5-4B & cohere & 107.0 & 108.0 & 137.0 & 117.3 & 123.0 & 109.0 & 156.0 & 129.3 \\
 & google & 113.0 & 116.0 & 119.0 & 116.0 & 121.0 & 108.0 & 145.0 & 124.7 \\
\cdashlinelr{1-10}
 & native & 196.0 & 180.5 & 186.0 & 187.5 & 188.0 & 178.0 & 202.0 & 189.3 \\
Qwen1.5-7B & cohere & 150.5 & 118.0 & 143.0 & 137.2 & 139.0 & 116.0 & 159.0 & 138.0 \\
 & google & 134.0 & 121.0 & 143.0 & 132.7 & 132.0 & 111.0 & 158.0 & 133.7 \\
\cdashlinelr{1-10}
 & native & 204.0 & 205.5 & 203.0 & 204.2 & 205.0 & 209.5 & 216.0 & 210.2 \\
Qwen1.5-14B & cohere & 142.0 & 151.0 & 165.0 & 152.7 & 150.0 & 129.0 & 158.0 & 145.7 \\
 & google & 137.0 & 132.0 & 151.0 & 140.0 & 141.0 & 134.0 & 147.0 & 140.7 \\
\bottomrule
\end{tabular}
\caption{Results for each model and each language on open-ended translated questions (GPT-4-Turbo judged).}
\label{tab:qa_translated_gpt4}
\end{table*}

\begin{table*}[ht]
\centering\small
\begin{tabular}{llcccGcccG}
\toprule
\multicolumn{1}{c}{\multirow[b]{2}{*}{Base model}} & \multicolumn{1}{c}{\multirow[b]{2}{*}{Data}} & \multicolumn{4}{c}{Monolingual} & \multicolumn{4}{c}{Multilingual} \\
\cmidrule(lr){3-6}\cmidrule(lr){7-10}
 & & es & ru & zh & \textit{average} & es & ru & zh & \textit{average} \\
\midrule
 & native & 184.0 & 167.0 & 129.0 & 160.0 & 180.0 & 168.0 & 138.0 & 162.0 \\
Llama2-7B & cohere & 174.5 & 162.0 & 147.0 & 161.2 & 183.0 & 166.0 & 150.0 & 166.3 \\
 & google & 179.0 & 165.0 & 148.0 & 164.0 & 178.0 & 162.0 & 137.0 & 159.0 \\
\cdashlinelr{1-10}
 & native & 168.0 & 155.0 & 126.0 & 149.7 & 170.0 & 150.0 & 149.0 & 156.3 \\
Gemma-2B & cohere & 166.0 & 154.0 & 152.0 & 157.3 & 166.0 & 160.0 & 151.0 & 159.0 \\
 & google & 162.0 & 159.0 & 144.0 & 155.0 & 168.0 & 153.0 & 157.0 & 159.3 \\
\cdashlinelr{1-10}
 & native & 194.0 & 190.0 & 152.0 & 178.7 & 190.0 & 181.0 & 166.0 & 179.0 \\
Gemma-7B & cohere & 171.0 & 174.0 & 168.0 & 171.0 & 178.0 & 161.0 & 161.0 & 166.7 \\
 & google & 186.0 & 172.0 & 170.0 & 176.0 & 181.0 & 172.0 & 164.0 & 172.3 \\
\cdashlinelr{1-10}
 & native & 131.0 & \phantom{0}96.0 & 131.0 & 119.3 & 121.0 & \phantom{0}99.0 & 138.0 & 119.3 \\
Qwen1.5-0.5B & cohere & 117.0 & \phantom{0}98.0 & 143.0 & 119.3 & 126.0 & \phantom{0}98.0 & 145.0 & 123.0 \\
 & google & 133.0 & 104.0 & 133.0 & 123.3 & 127.0 & 102.0 & 129.0 & 119.3 \\
\cdashlinelr{1-10}
 & native & 143.0 & 131.0 & 140.0 & 138.0 & 133.0 & 115.0 & 135.0 & 127.7 \\
Qwen1.5-1.8B & cohere & 147.0 & 128.0 & 152.0 & 142.3 & 145.0 & 121.0 & 159.0 & 141.7 \\
 & google & 163.0 & 125.0 & 145.0 & 144.3 & 148.0 & 124.0 & 152.0 & 141.3 \\
\cdashlinelr{1-10}
 & native & 179.0 & 160.0 & 155.0 & 164.7 & 169.0 & 149.0 & 170.0 & 162.7 \\
Qwen1.5-4B & cohere & 156.0 & 156.0 & 165.0 & 159.0 & 171.0 & 153.0 & 167.0 & 163.7 \\
 & google & 168.0 & 155.0 & 151.0 & 158.0 & 157.0 & 147.0 & 158.0 & 154.0 \\
\cdashlinelr{1-10}
 & native & 181.0 & 158.0 & 144.0 & 161.0 & 177.0 & 156.0 & 166.0 & 166.3 \\
Qwen1.5-7B & cohere & 178.0 & 159.0 & 154.0 & 163.7 & 183.0 & 148.0 & 160.0 & 163.7 \\
 & google & 174.0 & 153.0 & 161.0 & 162.7 & 172.0 & 141.0 & 168.0 & 160.3 \\
\cdashlinelr{1-10}
 & native & 183.0 & 172.0 & 156.0 & 170.3 & 182.0 & 173.0 & 169.0 & 174.7 \\
Qwen1.5-14B & cohere & 172.0 & 163.0 & 149.0 & 161.3 & 179.0 & 158.0 & 155.0 & 164.0 \\
 & google & 172.0 & 159.0 & 155.0 & 162.0 & 172.0 & 155.0 & 160.0 & 162.3 \\
\bottomrule
\end{tabular}
\caption{Results for each model and each language on open-ended translated questions (Command R+ judged).}
\label{tab:qa_translated_cmdr+}
\end{table*}

\begin{table*}[ht]
\centering\small
\begin{tabular}{llcccGcccG}
\toprule
\multicolumn{1}{c}{\multirow[b]{2}{*}{Base model}} & \multicolumn{1}{c}{\multirow[b]{2}{*}{Data}} & \multicolumn{4}{c}{Monolingual} & \multicolumn{4}{c}{Multilingual} \\
\cmidrule(lr){3-6}\cmidrule(lr){7-10}
 & & es & ru & zh & \textit{average} & es & ru & zh & \textit{average} \\
\midrule
 & native & 195.0 & 172.5 & 140.0 & 169.2 & 175.0 & 181.0 & 140.0 & 165.3 \\
Llama2-7B & cohere & 173.0 & 172.0 & 149.0 & 164.7 & 158.0 & 152.0 & 135.0 & 148.3 \\
 & google & 176.5 & 163.5 & 150.5 & 163.5 & 173.0 & 164.0 & 161.0 & 166.0 \\
\cdashlinelr{1-10}
 & native & 156.0 & 129.5 & 130.0 & 138.5 & 174.0 & 149.0 & 136.5 & 153.2 \\
Gemma-2B & cohere & 175.0 & 144.0 & 160.0 & 159.7 & 160.0 & 152.5 & 155.0 & 155.8 \\
 & google & 157.0 & 129.5 & 140.0 & 142.2 & 166.0 & 148.5 & 141.0 & 151.8 \\
\cdashlinelr{1-10}
 & native & 220.0 & 222.0 & 177.0 & 206.3 & 208.0 & 199.0 & 180.0 & 195.7 \\
Gemma-7B & cohere & 197.0 & 209.5 & 207.0 & 204.5 & 206.0 & 175.5 & 194.0 & 191.8 \\
 & google & 204.0 & 190.5 & 200.0 & 198.2 & 187.5 & 198.0 & 211.0 & 198.8 \\
\cdashlinelr{1-10}
 & native & \phantom{0}88.0 & \phantom{0}88.0 & 157.0 & 111.0 & \phantom{0}91.0 & \phantom{0}76.0 & 162.0 & 109.7 \\
Qwen1.5-0.5B & cohere & \phantom{0}91.5 & \phantom{0}91.0 & 145.5 & 109.3 & \phantom{0}78.0 & \phantom{0}70.0 & 156.0 & 101.3 \\
 & google & \phantom{0}93.0 & \phantom{0}66.0 & 142.0 & 100.3 & \phantom{0}91.0 & \phantom{0}72.0 & 157.0 & 106.7 \\
\cdashlinelr{1-10}
 & native & 121.0 & 107.0 & 174.0 & 134.0 & 118.0 & \phantom{0}88.0 & 156.0 & 120.7 \\
Qwen1.5-1.8B & cohere & 116.0 & 102.0 & 177.0 & 131.7 & 121.0 & 108.0 & 193.5 & 140.8 \\
 & google & 114.0 & \phantom{0}96.0 & 153.0 & 121.0 & 136.5 & 101.0 & 196.5 & 144.7 \\
\cdashlinelr{1-10}
 & native & 162.0 & 126.0 & 186.0 & 158.0 & 163.0 & 129.0 & 208.0 & 166.7 \\
Qwen1.5-4B & cohere & 152.0 & 141.0 & 173.0 & 155.3 & 168.5 & 140.5 & 208.0 & 172.3 \\
 & google & 146.0 & 137.0 & 191.0 & 158.0 & 161.0 & 133.0 & 213.0 & 169.0 \\
\cdashlinelr{1-10}
 & native & 191.0 & 179.0 & 176.0 & 182.0 & 201.0 & 139.0 & 186.0 & 175.3 \\
Qwen1.5-7B & cohere & 194.0 & 175.5 & 205.5 & 191.7 & 185.0 & 165.0 & 213.0 & 187.7 \\
 & google & 178.0 & 164.0 & 183.0 & 175.0 & 188.0 & 153.0 & 184.0 & 175.0 \\
\cdashlinelr{1-10}
 & native & 206.0 & 144.0 & 197.0 & 182.3 & 204.5 & 184.0 & 219.0 & 202.5 \\
Qwen1.5-14B & cohere & 194.0 & 206.5 & 216.0 & 205.5 & 211.0 & 187.5 & 236.0 & 211.5 \\
 & google & 177.0 & 167.0 & 222.5 & 188.8 & 197.0 & 182.0 & 212.0 & 197.0 \\
\bottomrule
\end{tabular}
\caption{Results for each model and each language on open-ended native questions (GPT-4-Turbo judged).}
\label{tab:qa_native_gpt4}
\end{table*}

\begin{table*}[ht]
\centering\small

\begin{tabular}{llcccGcccG}
\toprule
\multicolumn{1}{c}{\multirow[b]{2}{*}{Base model}} & \multicolumn{1}{c}{\multirow[b]{2}{*}{Data}} & \multicolumn{4}{c}{Monolingual} & \multicolumn{4}{c}{Multilingual} \\
\cmidrule(lr){3-6}\cmidrule(lr){7-10}
 & & es & ru & zh & \textit{average} & es & ru & zh & \textit{average} \\
\midrule
 & native & 194.0 & 165.0 & 131.0 & 163.3 & 184.0 & 172.0 & 144.0 & 166.7 \\
Llama2-7B & cohere & 173.0 & 160.0 & 145.0 & 159.3 & 184.0 & 163.0 & 149.0 & 165.3 \\
 & google & 184.0 & 161.0 & 153.0 & 166.0 & 179.0 & 158.0 & 142.0 & 159.7 \\
\cdashlinelr{1-10}
 & native & 168.0 & 151.0 & 128.0 & 149.0 & 179.0 & 152.0 & 155.0 & 162.0 \\
Gemma-2B & cohere & 176.0 & 152.0 & 155.0 & 161.0 & 178.0 & 155.0 & 151.0 & 161.3 \\
 & google & 178.0 & 153.0 & 140.0 & 157.0 & 179.0 & 162.0 & 151.0 & 164.0 \\
\cdashlinelr{1-10}
 & native & 201.5 & 187.0 & 148.0 & 178.8 & 202.0 & 181.0 & 169.0 & 184.0 \\
Gemma-7B & cohere & 183.0 & 172.0 & 162.0 & 172.3 & 180.0 & 162.0 & 168.0 & 170.0 \\
 & google & 190.0 & 180.0 & 174.0 & 181.3 & 182.0 & 175.0 & 170.0 & 175.7 \\
\cdashlinelr{1-10}
 & native & 133.0 & 103.0 & 153.0 & 129.7 & 136.0 & \phantom{0}96.0 & 150.0 & 127.3 \\
Qwen1.5-0.5B & cohere & 136.0 & \phantom{0}98.0 & 145.0 & 126.3 & 129.0 & \phantom{0}99.0 & 151.0 & 126.3 \\
 & google & 130.0 & \phantom{0}97.0 & 145.0 & 124.0 & 125.0 & 102.0 & 157.0 & 128.0 \\
\cdashlinelr{1-10}
 & native & 160.0 & 138.0 & 155.0 & 151.0 & 152.0 & 112.0 & 145.0 & 136.3 \\
Qwen1.5-1.8B & cohere & 151.0 & 121.0 & 170.0 & 147.3 & 149.0 & 118.0 & 167.0 & 144.7 \\
 & google & 156.0 & 128.0 & 166.0 & 150.0 & 161.0 & 125.0 & 163.0 & 149.7 \\
\cdashlinelr{1-10}
 & native & 180.0 & 152.0 & 161.0 & 164.3 & 181.0 & 149.0 & 171.0 & 167.0 \\
Qwen1.5-4B & cohere & 173.0 & 154.0 & 180.0 & 169.0 & 173.0 & 154.0 & 176.0 & 167.7 \\
 & google & 168.0 & 149.0 & 171.0 & 162.7 & 182.0 & 147.0 & 177.0 & 168.7 \\
\cdashlinelr{1-10}
 & native & 184.0 & 164.0 & 154.0 & 167.3 & 193.0 & 153.0 & 159.0 & 168.3 \\
Qwen1.5-7B & cohere & 184.0 & 154.0 & 160.0 & 166.0 & 184.0 & 146.0 & 173.0 & 167.7 \\
 & google & 181.0 & 155.0 & 166.0 & 167.3 & 182.0 & 149.0 & 164.0 & 165.0 \\
\cdashlinelr{1-10}
 & native & 191.0 & 160.0 & 152.0 & 167.7 & 192.0 & 167.0 & 169.0 & 176.0 \\
Qwen1.5-14B & cohere & 189.0 & 173.0 & 172.0 & 178.0 & 190.0 & 164.0 & 173.0 & 175.7 \\
 & google & 184.0 & 163.0 & 178.0 & 175.0 & 188.0 & 160.0 & 165.0 & 171.0 \\
\bottomrule
\end{tabular}
\caption{Results for each model and each language on open-ended native questions (Command R+ judged).}
\label{tab:qa_native_cmdr+}
\end{table*}

\end{document}